\documentclass[10pt,letterpaper]{article}
\usepackage[shownumpages,citingstyle=authoryear,bibliostyle=plainnat,bibfile=refs,boxrule=0.7pt]{brainlab}
\usepackage[T1]{fontenc}
\usepackage{textcomp}

\usepackage{amsmath,amsfonts,bm}

\def\eqref#1{equation~\ref{#1}}

\def\1{\bm{1}}

\DeclareMathAlphabet{\mathsfit}{\encodingdefault}{\sfdefault}{m}{sl}
\SetMathAlphabet{\mathsfit}{bold}{\encodingdefault}{\sfdefault}{bx}{n}

\usepackage{hyperref}
\usepackage{url}
\usepackage{booktabs}
\usepackage{xcolor}
\usepackage{wrapfig}

\usepackage{graphicx}

\usepackage{amsmath,amssymb}
\usepackage{microtype}
\usepackage{colortbl}
\usepackage{multirow}
\usepackage[section]{placeins}
\usepackage{float}
\usepackage{tcolorbox}
\tcbuselibrary{breakable}

\definecolor{routeplum}{HTML}{854C65}
\definecolor{routesunset}{HTML}{F0C987}
\definecolor{routepapaya}{HTML}{FBEDD6}
\definecolor{routefloral}{HTML}{FAF8F1}
\definecolor{routeblue}{HTML}{9DE1FC}
\colorlet{HWPrimary}{braincolor}
\hypersetup{colorlinks=true,allcolors=braincolor}

\newcommand{\ce}{\mathrm{CE}}

\definecolor{meanfill}{HTML}{E8F0F7}

\newtcolorbox{takeawaybox}{
  colback=white,
  colframe=HWPrimary!50!white,
  boxrule=0.8pt,
  arc=2mm,
  left=2mm,
  right=2mm,
  top=1mm,
  bottom=1mm,
  before skip=3pt,
  after skip=1pt,
  breakable
}

\renewcommand{\brainprintbibliography}{%
  \clearpage
  \begingroup\small\setlength{\bibsep}{2pt plus 1pt}
  \bibliographystyle{plainnat}
  \bibliography{refs}
  \endgroup
}
\setbrainmeta{
  title={\fontsize{17}{20}\selectfont
     Multi-Stream Residual Computation \\ in Diffusion Language Models
  },
  authors={%
    \normalfont\mdseries\normalsize
    \setlength{\tabcolsep}{4pt}
    \begin{tabular*}{\linewidth}{@{\extracolsep{\fill}}cc@{}}
      \makebox[0.48\linewidth][c]{\textbf{Ekaterina Alimaskina}} &
      \makebox[0.48\linewidth][c]{\textbf{Suren Kazarian}}\\[2pt]
      {\small BRAIn Lab} & {\small BRAIn Lab}\\[2pt]
      {\small Yandex Research} & \\[1em]
      \textbf{Gleb Molodtsov} & \textbf{Aleksandr Beznosikov}\\[2pt]
      {\small BRAIn Lab} & {\small BRAIn Lab}
    \end{tabular*}
  },
  affiliations={},
  code={\vspace{-2mm}{\small
    \url{https://github.com/brain-lab-research/diffusion-residual-streams}\vspace{4mm}
  }},
  abstract={
  Diffusion language models reconstruct masked tokens in parallel, offering an alternative to autoregressive generation. Yet their Transformer backbones typically carry visible evidence and evolving masked predictions through a single residual stream. We equip masked diffusion models with multiple residual streams that persist through depth and adaptive routing between them, then investigate how they support these complementary roles. On the pretraining task, two streams reduce reconstruction perplexity by 5.71\% on validation data and 5.13\% on average across external corpora compared with standard residual connections. We observe that, without predefined roles, visible and masked positions learn distinct routing patterns. Causal interventions show that this distinction organizes read and write routing, while token-specific write strength remains important within each role. Guided by these findings, a model trained with role-conditioned routing recovers half the validation gain from adding a second stream. State interventions explain the benefit of the additional stream: it preserves early token representations for later reuse while evidence accumulates at masked positions. This mechanism recurs at a larger model scale, showing how persistent streams support context preservation alongside prediction refinement.
}
}

\begin{document}
\maketitlebox

\section{Introduction}

Diffusion language models have recently emerged as an alternative to autoregressive language modeling \citep{austin2021structured,hoogeboom2021argmax,sahoo2024simple,nie2025llada,ye2025dream}. Rather than generating text token by token, they predict many missing positions in parallel and refine uncertain predictions over successive steps. Reconstruction can use context from both sides, supporting flexible generation orders and naturally accommodating text infilling.

The two generation schemes also differ in what they require of each token position. Under next-token training, autoregressive positions share a largely uniform prediction task: each uses its prefix to predict the following token. Masked diffusion combines two complementary roles within the same denoising pass. Visible positions carry observed tokens that supply evidence for reconstruction, while masked positions must gather contextual information to form and refine predictions. These roles coexist in every partially corrupted sequence, so the backbone must accommodate observed evidence and evolving predictions across positions and through depth.

Despite these different roles, most masked diffusion models inherit the standard Transformer residual pathway. Attention and MLP branches can respond to the corruption mask, but at each position they read from and write to a single persistent state. We ask whether learned routes through depth can better support these complementary roles.

Work beyond text diffusion shows that residual pathways can be made learnable. ReZero and DenseFormer regulate updates or combine information across depth, while Hyper-Connections introduce multiple persistent streams with learned read, mixing, and write patterns \citep{bachlechner2021rezero,pagliardini2024denseformer,zhu2025hyperconnections,xie2025mhc}. For masked diffusion, this flexibility offers a way to preserve useful evidence while refining predictions, with routing that can adapt to the different roles of visible and masked positions.

We first introduce token-conditioned update strength within a single residual stream. This improves diffusion modeling and provides a learned baseline for evaluating the benefit of multiple persistent states. Two streams capture most of the further gain, so we focus on how the model uses this added capacity. Causal interventions show that the visible/masked distinction organizes read and write-destination routing, while token-specific write volume remains important within each role.

\begin{figure}[t]
    \centering
    \includegraphics[width=\linewidth]{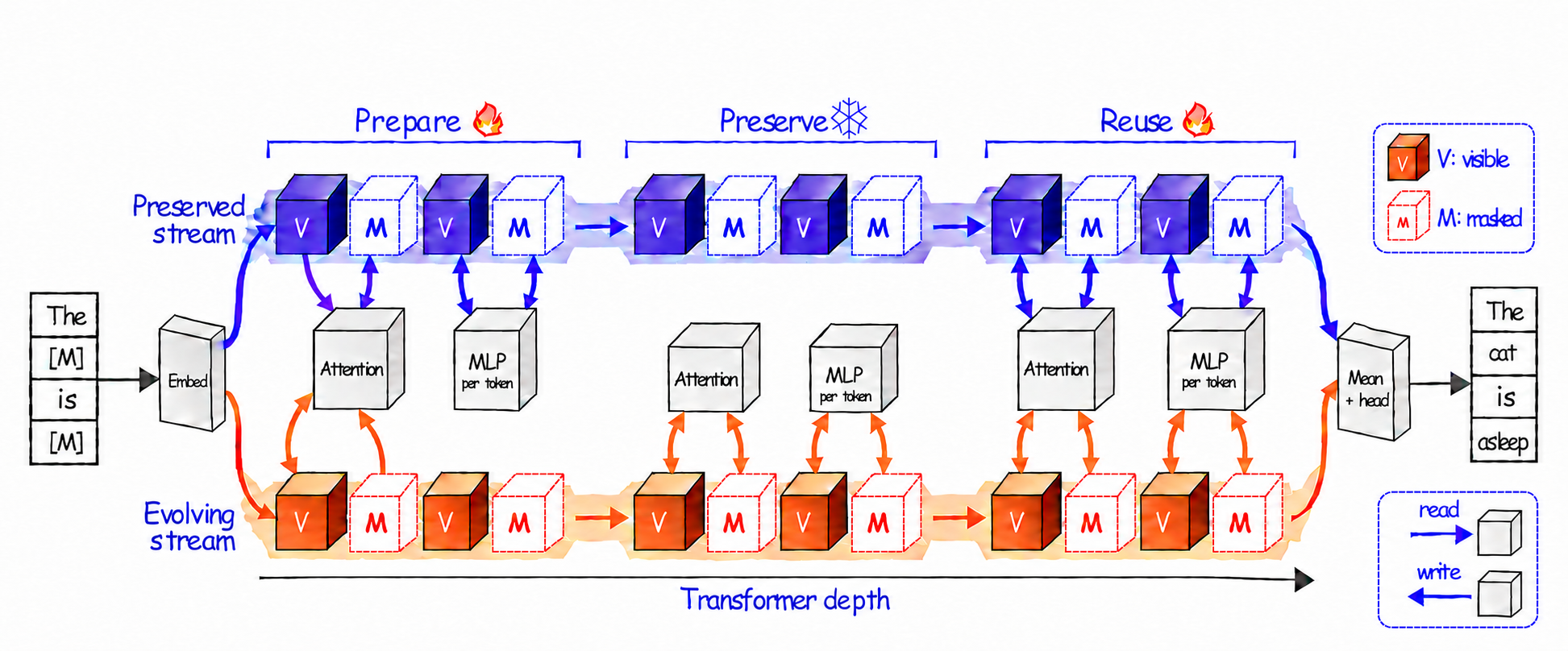}
    \caption{
        Prepare--preserve--reuse within a denoising pass.
        Early blocks prepare position-specific states, one stream
        preserves them while the other accumulates evidence,
        and late blocks reuse the retained information.
        Arrows show selected read and write routes;
        weak connections are omitted.
    }
    \label{fig:stream-routing}
\end{figure}

Stream-state interventions show how the extra state supports denoising: it keeps early information available while the model gathers evidence for masked predictions. One stream retains position-specific states formed in early blocks through middle depth, as visible-source influence accumulates at masked targets. Late layers then increase reading from this stream and reuse the retained visible state. This preservation--transfer--reuse sequence recurs at a larger model scale, while preservation followed by renewed late reading also extends across stream counts. These findings motivate residual pathways that adapt to token roles and coordinate evidence preservation with prediction refinement.

\paragraph{Contributions.}
Our main contributions are threefold:

\begin{itemize}
    \item \textbf{Residual expansion.} We establish the benefits of learned residual transport in masked diffusion. Two persistent streams capture 86.6\% of the largest additional validation-CE gain beyond learned single-stream control in the stream-count sweep and achieve the largest mean masked-reconstruction perplexity reduction across four external corpora.

    \item \textbf{Role specialization.} We separate write destination from write volume and show through controlled interventions that masked/visible roles organize read and write-destination routing, while token-specific write volume matters for visible evidence early and masked predictions late. These findings also guide a role-factorized interface trained from scratch.

    \item \textbf{State reuse.} We uncover a preservation--transfer--reuse mechanism linking persistent states to denoising computation: one stream preserves early position-specific states through middle depth while visible-source influence accumulates at masked targets, and late layers reuse the retained visible state. Its functional stages recur at a larger model scale. Models with two, three, and four streams also preserve early states and resume reading them in late layers.
\end{itemize}

\newpage

\section{Related Work}

\paragraph{Text diffusion language models.}
Discrete diffusion extends corruption and denoising to categorical sequences \citep{austin2021structured,hoogeboom2021argmax,lou2024discrete}. Masked diffusion develops this approach for language modeling, training models to reconstruct missing tokens from partially observed sequences \citep{sahoo2024simple}. Recent systems demonstrate progress in training objectives, decoding, and scale \citep{nie2025llada,ye2025dream}. These advances establish denoising as a practical alternative to autoregressive generation, while generally retaining the standard Transformer residual pathway. We investigate this architectural choice: whether multiple persistent states and learned routes through depth can better support the computation performed within a denoising pass.
\vspace{-4mm}
\paragraph{Mask-dependent computation in diffusion models.}
Recent work shows that token mask state affects both context processing and internal representations. Appended mask tokens can act as distractors and impair context comprehension \citep{piskorz2026masks}. Masked and unmasked positions exhibit different activation distributions, motivating state-specific transformations for quantization \citep{yan2026starquant}. Reusing hidden states across denoising steps also benefits from distinguishing unchanged, still-masked, and newly revealed tokens \citep{miele2026fixedpoint}. These findings motivate treating mask state as a computational variable. We ask how the complementary roles of visible evidence and masked predictions organize the processing, preservation, and reuse of information through network depth.
\vspace{-4mm}
\paragraph{Residual architectures outside text diffusion.}
Residual redesigns offer forms of control over information flow. ReZero introduces zero-initialized learnable gates on residual branches \citep{bachlechner2021rezero}, while DeepNorm combines residual scaling with depth-dependent initialization \citep{wang2022deepnet}. DenseFormer learns weighted combinations of current and earlier representations \citep{pagliardini2024denseformer}, while Attention Residuals aggregates representations across depth using input-dependent softmax attention \citep{kimi2026attentionresiduals}. They regulate updates or retrieve information from layer history. In masked diffusion, visible positions provide evidence, whereas masked positions construct predictions from it. This distinction motivates asking whether residual routing should adapt to a token’s role and current state, beyond update scaling and retrieval from layer history.

Hyper-Connections provide an interface through parallel streams and learned maps governing how sublayers read, mix, and update them, with token-dependent connection weights in the dynamic variant \citep{zhu2025hyperconnections}. mHC constrains mixing toward doubly stochastic matrices via Sinkhorn--Knopp normalization \citep{xie2025mhc}, while mHC-lite constructs them exactly as convex combinations of permutation matrices, avoiding iterative normalization \citep{yang2026mhclite}. We extend this interface to text diffusion, establishing its pretraining benefits and revealing how denoising organizes residual information flow without prescribed routes for masked or visible positions.
\vspace{-4mm}
\paragraph{Mechanistic analyses of multi-stream residual pathways.}
Analyses of autoregressive HC and mHC models report asymmetric stream reliance, concentrated read/write routing, and near-identity mixing \citep{peng2026ablate,alimaskina2026streamcollapse,zhao2026mhcstreams}. These findings motivate going beyond aggregate performance to understand how learned interfaces use the additional streams. We study corruption-role-conditioned routing, separate write destination from volume, and trace source-dependent prediction recovery alongside the preservation of early states.

\section{Decomposing Residual Transport}
\label{sec:interface}
\paragraph{Multi-stream residual interface.}
Following the mHC notation, let
$\boldsymbol{x}_{\ell}\in\mathbb{R}^{n\times d}$ contain the residual
streams of one token entering sublayer $\ell$, with the token index omitted. For attention, $\mathcal{F}_{\ell}$ jointly consumes the read representations of all positions. The interface updates the streams as
\begin{equation}
\boldsymbol{x}_{\ell+1}
=
\mathcal{H}_{\ell}^{\mathrm{res}}\boldsymbol{x}_{\ell}
+
\mathcal{H}_{\ell}^{\mathrm{post}\,\top}
\mathcal{F}_{\ell}
\!\left(
\mathcal{H}_{\ell}^{\mathrm{pre}}\boldsymbol{x}_{\ell}
\right).
\label{eq:residual-transport}
\vspace{-1mm}
\end{equation}
The token-conditioned mappings
$\mathcal{H}_{\ell}^{\mathrm{pre}}$,
$\mathcal{H}_{\ell}^{\mathrm{res}}$, and
$\mathcal{H}_{\ell}^{\mathrm{post}}$ respectively determine what the
sublayer reads, how existing streams are carried forward, and how its output
is written back. In mHC-lite,
$\mathcal{H}_{\ell}^{\mathrm{res}}$ is exactly doubly stochastic and
therefore preserves the mean across streams. Streams are initialized by
replication and averaged before the final output head. Blocks are indexed from zero; $a_k$ and $m_k$ denote their attention and MLP interfaces.
\vspace{-3mm}
\paragraph{Separating write destination from volume.}
The post-mapping controls both where a sublayer update is written and how strongly it enters the residual streams. Its entries conflate these two decisions; normalizing them separates allocation from strength:
\vspace{-1mm}
\begin{equation}
q_{\ell} = \frac{\mathcal{H}_{\ell}^{\mathrm{post}}}{\mathcal{H}_{\ell}^{\mathrm{post}}\mathbf{1}_{n}}, \qquad v_{\ell} = \frac{1}{n}\mathcal{H}_{\ell}^{\mathrm{post}}\mathbf{1}_{n}.
\label{eq:write-decomposition}
\end{equation}
\vspace{-1mm}

Here $q_{\ell}\mathbf{1}_{n}=1$ is the \emph{write destination}, while $v_{\ell}$ is the \emph{write volume}. Equivalently, $\mathcal{H}_{\ell}^{\mathrm{post}}=nv_{\ell}q_{\ell}$: $q_{\ell}$ determines how the update is divided among streams, and $v_{\ell}$ scales that allocation.

Because $\mathcal{H}_{\ell}^{\mathrm{res}}$ is doubly stochastic, it transports the existing streams without changing their mean. Averaging Equation~\ref{eq:residual-transport} over streams therefore gives
\begin{equation}
\bar{\boldsymbol{x}}_{\ell+1} = \bar{\boldsymbol{x}}_{\ell} + v_{\ell}\mathcal{F}_{\ell}\!\left(\mathcal{H}_{\ell}^{\mathrm{pre}}\boldsymbol{x}_{\ell}\right),
\end{equation}
\vspace{-3mm}

where $\bar{\boldsymbol{x}}_{\ell}$ is the mean across streams. Thus $v_{\ell}$ sets the coefficient of the branch update in the mean state: it is a token-dependent counterpart of the unit update coefficient in a standard residual connection. The update norm also depends on the branch output. The destination $q_{\ell}$ determines which individual streams receive that update, shaping the states available to later sublayers.

We analyze the HC-family post-mapping \citep{zhu2025hyperconnections,xie2025mhc,yang2026mhclite} via normalized write destination and overall write strength, enabling separate interventions on these two coordinates. Our results show this distinction matters: masked versus visible role largely organizes reading and write destination, while write volume is most sensitive to within-role averaging. Replacing token-dependent write volume also degrades autoregressive models (Appendix~\ref{app:ar}), suggesting that token-dependent update strength is a general component of learned residual transport.

\vspace{2mm}

\begin{takeawaybox}
\textbf{Takeaway.} The residual interface has four controls: reading selects what a sublayer uses, mixing redistributes existing states across streams, write destination allocates the new update, and write volume controls its overall strength.
\end{takeawaybox}

\section{How Many Residual Streams Does Diffusion Need?}
\label{sec:width}

\begin{wrapfigure}{r}{0.44\textwidth}
  \centering
  \includegraphics[width=\linewidth]{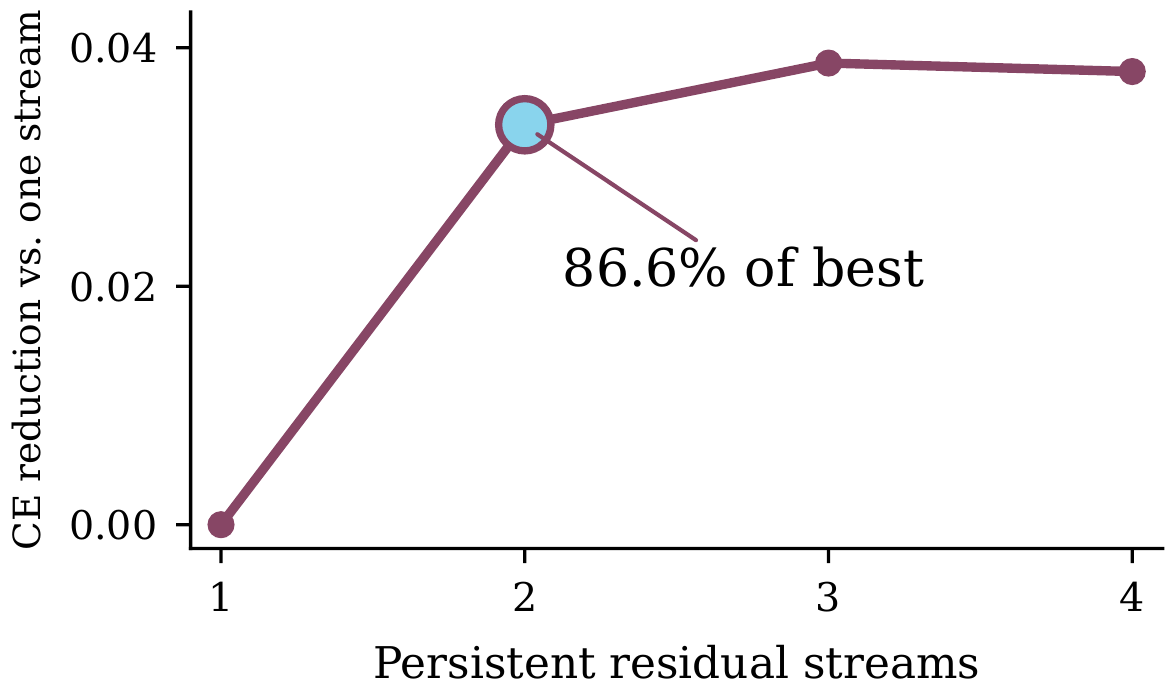}
  \caption{\textbf{Two streams capture most gain from additional streams.}
  Validation-CE reduction relative to learned $n=1$.
  The $n=2$ model captures 86.6\% of the largest additional gain
  in the matched stream-count sweep.}
  \label{fig:width}
\end{wrapfigure}

Before analyzing how diffusion uses residual streams, we ask how many are useful. We compare mHC-lite variants ($n=1,2,3,4$) with a standard-residual baseline using a 16-block Qwen-style backbone \citep{qwen2025qwen25}, approximately 337M parameters, and the Dream tokenizer \citep{ye2025dream}. All models are trained on the same ordered 3.5B tokens from FineWeb-Edu \citep{penedo2024fineweb}, with the backbone, token budget, and learning-rate schedule held fixed. The learned $n=1$ interface provides token-dependent update strength, while wider variants add routing across persistent streams. This width sweep guides the choice of model for the mechanistic analysis. Full training and evaluation details appear in Appendices~\ref{app:protocol} and~\ref{app:width}.

Even one learned stream improves over the standard residual, reducing validation reconstruction perplexity by $2.50\%$. Two streams bring this reduction to $5.71\%$, capturing $86.6\%$ of the best additional CE gain beyond learned $n=1$ (Figure~\ref{fig:width}). Adding a third or fourth stream yields only a further $0.52\%$ or $0.45\%$ reduction in perplexity relative to $n=2$. Since most of the benefit is already achieved with two streams, we use $n=2$ for the mechanistic analysis. At 1B, two streams reduce validation reconstruction perplexity by $3.47\%$ versus the matched standard residual (Appendix~\ref{app:scale-transfer}).

\begin{wraptable}{r}{0.55\textwidth}
  \centering
  \caption{\textbf{Two streams transfer best on average.}
  Masked-reconstruction perplexity reduction (\%) over the
  standard residual. Higher is better; column bests are bold.}
  \label{tab:width-main}
  \small
  \renewcommand{\arraystretch}{1.12}
  \setlength{\tabcolsep}{2.3pt}
  \begin{tabular}{@{}lcccc>{\columncolor{meanfill}}c@{}}
    \toprule
    Interface & OWT & WT103 & LAMBADA & C4 & \textbf{Mean} \\
    \midrule
    Standard residual & 0.00 & 0.00 & 0.00 & 0.00 & 0.00 \\
    $n=1$ & 1.44 & 1.17 & $-0.15$ & 1.77 & 1.06 \\
    $\boldsymbol{n=2}$ & 4.61 & \textbf{7.61} & \textbf{3.02} & 5.29 & \textbf{5.13} \\
    $n=3$ & \textbf{5.51} & 6.15 & 2.27 & 5.17 & 4.78 \\
    $n=4$ & 5.14 & 3.30 & 2.67 & \textbf{5.39} & 4.13 \\
    \bottomrule
  \end{tabular}
\end{wraptable}

To test whether the benefit of two streams extends beyond FineWeb-Edu, we evaluate all checkpoints on OpenWebText, WikiText-103, LAMBADA, and C4 using identical examples and masks at three corruption levels. All multi-stream variants improve over the standard-residual baseline on every corpus. Although wider models lead on individual corpora, $n=2$ achieves the largest mean perplexity reduction, at $5.13\%$ (Table~\ref{tab:width-main}). Full details and paired bootstrap intervals appear in Appendix~\ref{app:width}. 

\vspace{2mm}

\begin{takeawaybox}
\textbf{Takeaway.} Two streams capture most of the in-domain benefit of residual expansion and achieve the strongest mean transfer across four corpora. We therefore use $n=2$ to trace how denoising uses the learned interface.
\end{takeawaybox}

\section{The Corruption Mask Organizes Residual Routing}
\label{sec:roles}

The second stream improves performance beyond token-dependent update strength, suggesting that routing across persistent states also matters. We therefore ask what guides this routing. In masked diffusion, the corruption mask distinguishes positions that provide observed evidence from those that predict missing content. Our interventions test which routing decisions follow this role distinction and which require finer token-specific information.

\vspace{-2mm}

\paragraph{Causal interventions.}
We study the $n=2$ model at $t=0.5$ with 50\% masking, where both token roles are well represented. Each intervention targets one control -- read, mixing, write destination, or write volume -- throughout the network. The remaining controls are computed normally from the intervened states. At each attention and MLP interface, we test three replacements: a \emph{global mean} shared by all tokens, separate \emph{role means} for masked and visible tokens, and a \emph{role swap} that exchanges these two means. We also test joint replacement of all four controls. Calibration uses sequences disjoint from evaluation, and roles are identified from the corrupted input. Appendix~\ref{app:protocol} specifies how stream-wise write gains are reconstructed for destination, volume, and joint interventions.

\begin{figure}[H]
\centering
\includegraphics[width=\linewidth]{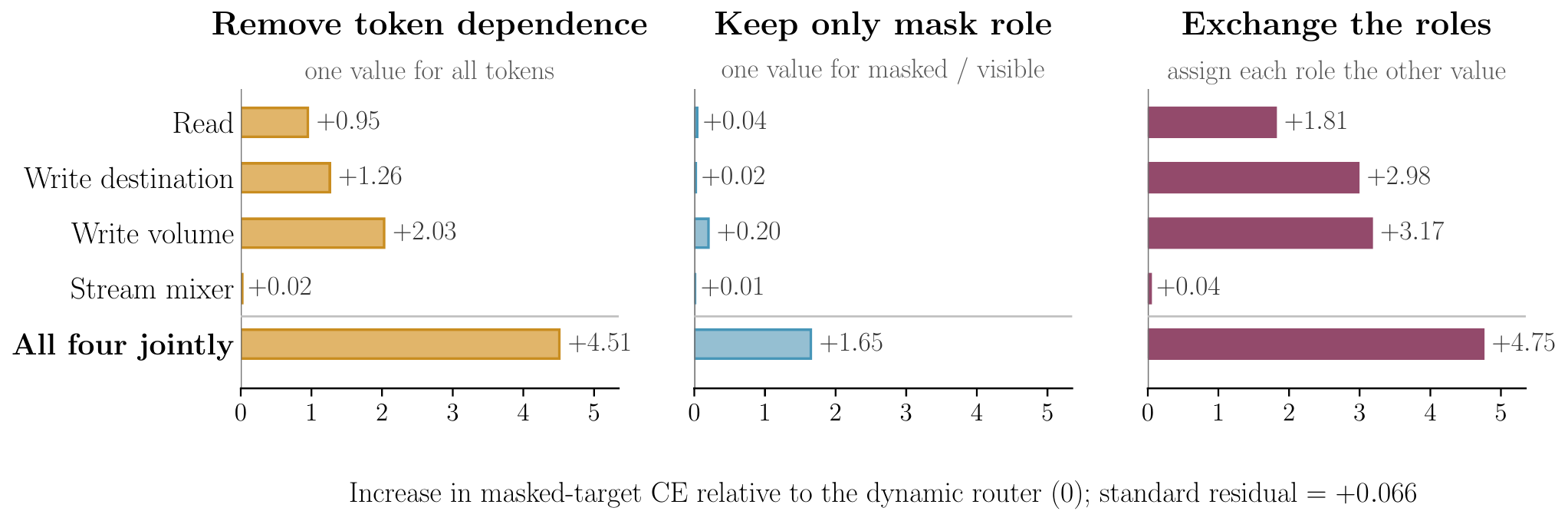}
\caption{\textbf{Masked and visible tokens use distinct residual routes.} Role-conditioned means remove most degradation from global averaging of read and write destination. Swapping either setting performs worse than the standard residual. A shared mixer nearly preserves performance, while write volume retains substantial within-role dependence. Lower masked-target CE is better. Individual rows replace one control, leaving the others state-dependent; the joint row replaces all four. Estimates and confidence intervals appear in Appendix~\ref{app:role-table}.}
\label{fig:role-ladder}
\end{figure}

Figure~\ref{fig:role-ladder} shows that collapsing read/write destination to one mean substantially degrades performance. Separate means for masked and visible positions largely remove this degradation. For read and write destination, role-conditioned means remove 95.7\% and 98.6\% of the CE increase caused by global averaging, respectively, measured relative to the intact model. 

This identifies the corruption mask as a major organizing factor for read and write-destination routing. Swapping the two role means sharply degrades denoising, showing that these routes are functionally specialized and must remain matched to token role. Mixing behaves differently: using a single mean at each interface causes little degradation, indicating that a largely shared mixing configuration supports denoising. The benefit of role-conditioned read and write-destination means recurs in the larger 1B model (Appendix~\ref{app:scale-allocation}).

Write volume remains sensitive even to role-conditioned averaging, showing that token-specific update strength matters within each role. Applying role means to all four controls at once is substantially more disruptive than replacing any one individually: joint replacement raises CE by $1.6453$ (Appendix~\ref{app:role-table}). We therefore test role-conditioned routing in a factorized interface trained from scratch, allowing its controls to adapt together.

\vspace{-2mm}

\paragraph{Training a role-factorized interface.}
Our role-factorized two-stream model, termed Factorized Residual Routing (FRR), learns separate read and write-destination settings for masked and visible tokens at each attention and MLP interface. Write volume remains token-dependent, while stream mixing is fixed to the identity. Write gains factorize into volume and normalized destination, retaining the full router’s total write-volume range. The backbone, data order, corruption masks, and optimization schedule match the other learned-interface models (Appendix~\ref{app:frr}).

FRR reaches a validation CE of $3.5634$, improving over learned $n=1$ ($3.5802$) and recovering $50.2\%$ of the gain achieved by full $n=2$ ($3.5467$). Thus, the masked/visible distinction can guide the learning of useful read and write routes: even with identity mixing, role-conditioned routes and token-dependent write strength allow the model to benefit from a second persistent state. Full $n=2$ provides a further $0.0167$ CE improvement.

\vspace{2mm}

\begin{takeawaybox}
\textbf{Takeaway.} The masked/visible distinction guides read and write routing, while token-specific write strength remains consequential within each role. FRR turns this division into a trained architecture that recovers half the gain from a second stream. 
\end{takeawaybox}

\section{Write Volume Matters for Visible Tokens Early and Masked Tokens Late}
\label{sec:volume}

Masked/visible role organizes read and write-destination routing, while write volume carries functionally important variation within each role. Tokens with the same role can follow similar routes while receiving updates of different strength. At $t=0.5$, replacing write volume with its masked/visible mean increases CE by $0.200$, compared with $0.041$ for read, $0.017$ for write destination, and $0.006$ for mixing. We therefore ask whether denoising depends on matching each volume to the token state that produced it, and how this dependence changes across depth and corruption levels.

\vspace{-3mm}

\paragraph{Causal interventions.} We first test whether each volume must remain attached to the token state that produced it. A \emph{role-mean replacement} assigns the calibrated mean for the token's role and interface wherever the proportional replacement satisfies the gain bounds, removing within-role variation at those positions. A \emph{within-role shuffle} exchanges current volumes between random disjoint pairs of same-role positions within each sequence and interface. An exchange is applied only when both directions satisfy the gain bounds, preserving the exact set of volume values within the group. Both interventions preserve each affected token's current normalized write destination. The reconstruction and acceptance rules are specified in Appendix~\ref{app:protocol}.

We then locate this dependence through depth by applying the role-mean replacement to one block at a time. Finally, we repeat the full-network replacement across corruption levels to determine when token-specific volume is most useful. Read and mixing remain dynamic; calibration uses disjoint sequences, and effects are measured by the increase in masked-target CE.

\begin{figure}[H]
\centering
\includegraphics[width=\linewidth]{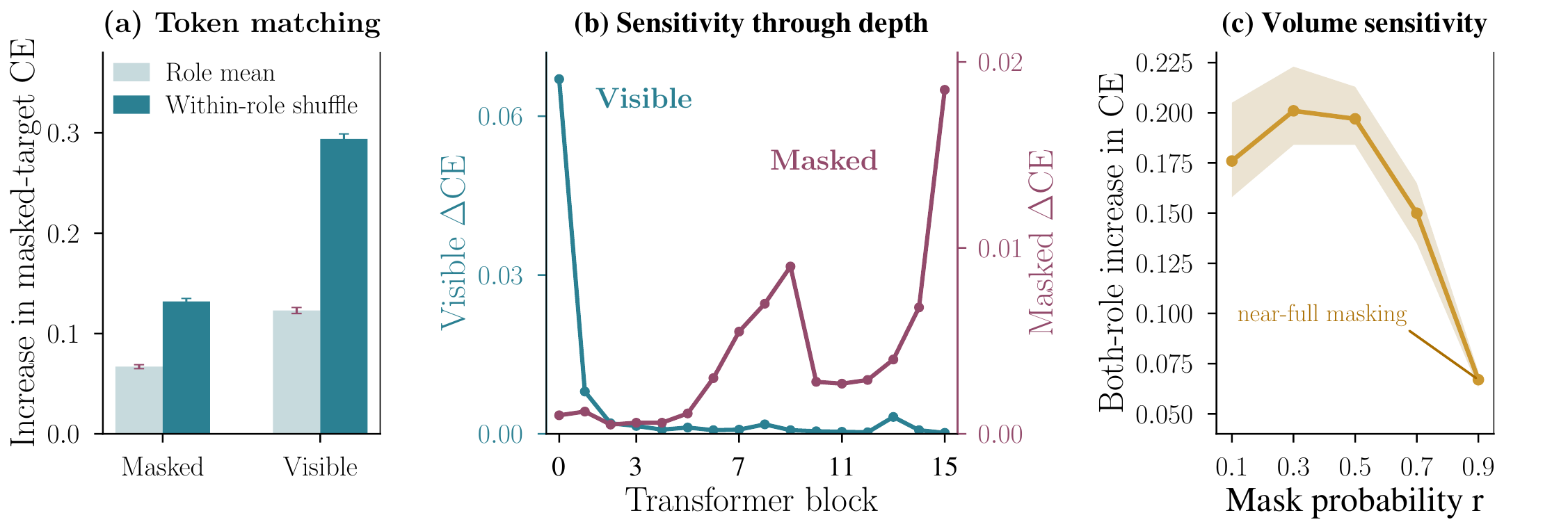}
\caption{\textbf{The functional role of token-specific write volume shifts through depth.}
(a) At $t=0.5$, within-role shuffling is more damaging than role-mean replacement while preserving the current volume multiset within each sequence and role at every intervened interface.
(b) Single-block interventions reveal an early-visible / late-masked sensitivity profile; effects use separate zero-anchored axes.
(c) Sensitivity to within-role volume variation decreases near full masking.
Error bars and bands show paired 95\% sequence-bootstrap intervals; $n=1{,}024$ in (a--b) and 128 per corruption level in (c).}
\label{fig:volume-schedule}
\end{figure}

\vspace{-4mm}

\paragraph{Write volume is matched to token state.}
Within-role shuffling is more damaging than role-mean replacement for both roles (Figure~\ref{fig:volume-schedule}a). For masked positions, shuffling increases CE by $0.132$, compared with $0.068$ under averaging; for visible positions, the corresponding increases are $0.294$ and $0.124$. Accepted exchanges retain the original write strengths but assign them to different tokens of the same role. The CE increase under this distribution-preserving exchange establishes the functional importance of matching write strength to token state within each role.

\vspace{-4mm}

\paragraph{Write-volume dependence shifts through depth.}
Single-block interventions locate this token--volume dependence through depth (Figure~\ref{fig:volume-schedule}b). In the first block, replacing visible-token volume increases CE by $+0.0673$, while replacing masked-token volume has an effect of only $+0.0015$. Visible-token sensitivity peaks in the first block; masked-token sensitivity peaks near the readout.

These interventions show that denoising depends most on visible-token write volume early and masked-token write volume late. Since volume scales the branch update, this depth profile links early evidence integration to later prediction refinement within a single denoising pass. The zero visible effect in the final block follows from the absence of subsequent attention. Across interior blocks, visible-token sensitivity declines sharply after the first block, while masked-token sensitivity follows a non-monotonic profile that peaks near the readout. Token-specific volume assignment and the early-visible / late-masked endpoint contrast also recur
at 1B (Appendix~\ref{app:scale-allocation}).

\vspace{-4mm}

\paragraph{Token-specific volume is most consequential under partial observation.}
Finally, we test how the importance of token-specific volume changes with corruption level (Figure~\ref{fig:volume-schedule}c). Replacing volume with role means increases CE by $0.177$--$0.201$ at 10--50\% masking, but by only $0.067$ at 90\% masking, when few visible tokens remain. The largest measured CE cost thus occurs with substantial visible context. Together with the depth interventions, this points to a role for write volume in regulating how the model uses observed evidence to predict missing content.

\begin{takeawaybox}
\textbf{Takeaway.} Denoising depends on matching write strength to token state: within-role shuffling degrades predictions despite preserving volume distributions. Sensitivity peaks for visible tokens in the first block and masked tokens near the readout, and is greatest with substantial visible context. Write volume helps coordinate evidence integration and prediction refinement.
\end{takeawaybox}

\section{Residual Streams Preserve Early States While Accumulating Evidence}
\label{sec:circuit}

\vspace{-1mm}

We now trace the information carried by the two streams through depth in the 337M model, then compare their functional stages with 1B (Appendix~\ref{app:scale-transfer}). In this 337M model, stream~0 is the \emph{evolving stream}, $s_{\mathrm E}$, and stream~1 the \emph{preserved stream}, $s_{\mathrm P}$. These names describe their middle-depth functions: $s_{\mathrm E}$ receives ongoing updates, while $s_{\mathrm P}$ stays close to its early post-$m_0$ state.

\begin{figure*}[t]
\centering
\includegraphics[width=0.875\textwidth]{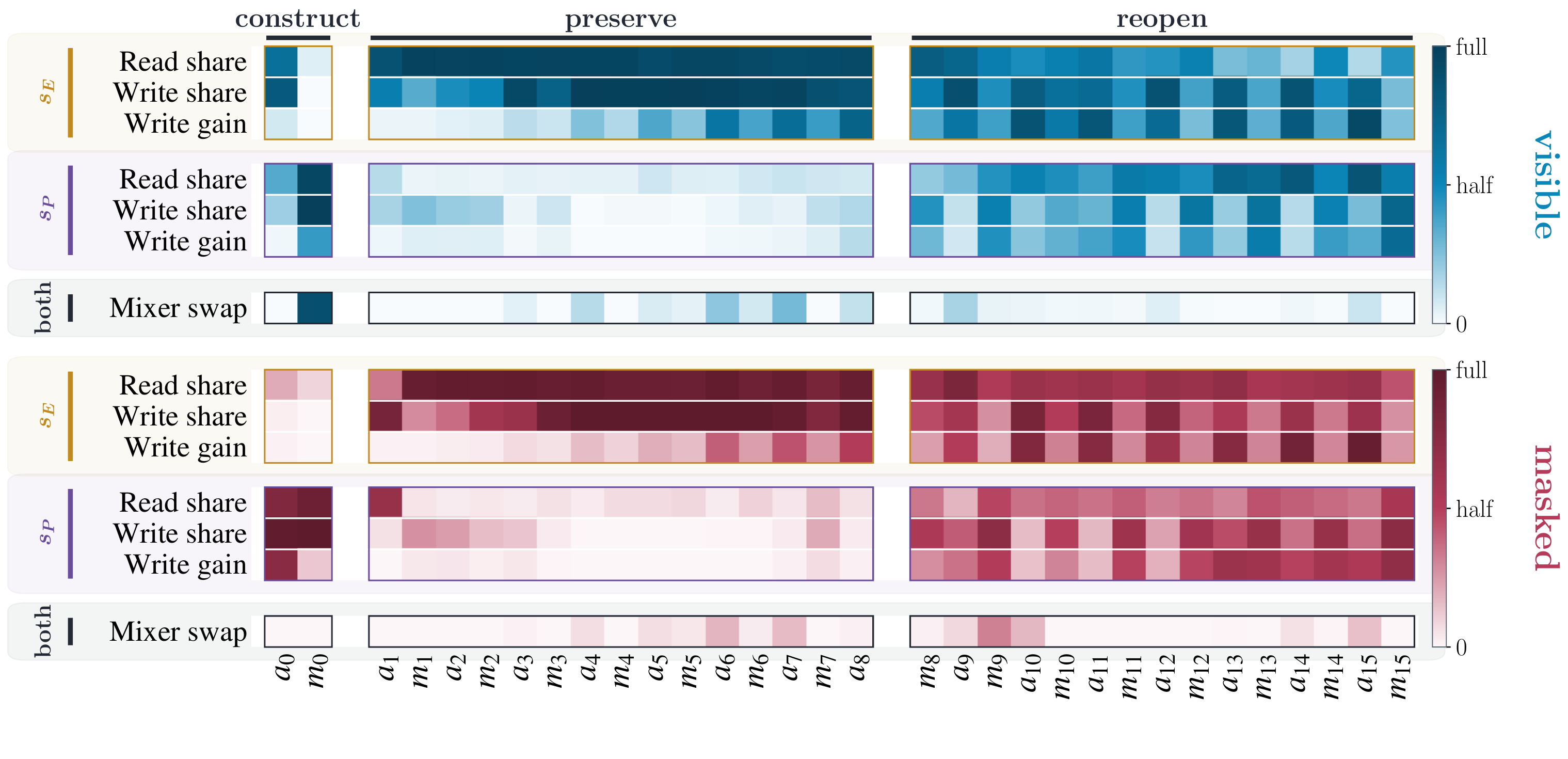}
\caption{\textbf{The preserved stream has little read/write activity at middle depth and reopens late.} Columns show the $32$ attention/MLP interfaces ($a_k$, $m_k$); rows show per-stream read share, write share $\overline{g}_{\ell i}/\sum_j\overline{g}_{\ell j}$, and role-mean physical write gain $\overline{g}_{\ell i}$. Mixer off-diagonal mass is shown once per role, with visible positions above and masked positions below. Scales are $0$--$1$ for shares, $0$--$2$ for gains, and $0$--$0.25$ for mixer mass. Across $a_1$--$a_8$, $s_{\mathrm P}$ receives $6\%$ of read mass and $3\%$ of write gain.}
\label{fig:circuit-atlas}
\end{figure*}

\vspace{-5mm}

\paragraph{Early blocks prepare position-specific states.}
In the 337M model, this preparation is concentrated in the first block. The first attention interface leaves $s_{\mathrm P}$ largely unchanged at visible positions but writes context into it at masked positions. Its write shares computed from role-mean gains for $(s_{\mathrm E},s_{\mathrm P})$ are $(0.88,0.12)$ and $(0.02,0.98)$, respectively (Figure~\ref{fig:circuit-atlas}). The first MLP, $m_0$, then reads mainly from $s_{\mathrm P}$ and adds its update mainly back to it. Together, these operations form an early state from the information initially available at each position: the observed token itself at visible positions and the context gathered by the first attention branch at masked positions. Meanwhile, $s_{\mathrm E}$ receives the attention update at visible positions but remains closer to the mask input at masked positions, providing different starting states for the ongoing computation.

The interventions test whether the next MLP needs this role-dependent input. At masked positions, removing the attention update sharply increases CE, but restoring only its contribution to the following MLP input removes $98.2\%$ of that increase. At visible positions, redirecting the attention output into $s_{\mathrm P}$ instead of $s_{\mathrm E}$ increases CE by $3.04$. These results show that the first block relies on role-dependent routing to determine what information the MLP receives when forming $s_{\mathrm P}$.

\vspace{-4mm}

\paragraph{An early state remains useful while changing little.}
Through the middle layers, attention and MLP branches predominantly read from and write to $s_{\mathrm E}$, while $s_{\mathrm P}$ receives little read or write activity and mixing remains nearly diagonal, so little of the preserved state is mixed into the active stream (Figure~\ref{fig:circuit-atlas}). Its similarity to its own post-$m_0$ state also remains high (Appendix~\ref{app:state-preservation}). We test whether this early state can still replace the current state, and whether it must belong to the same position.

At $a_8$, replacing $s_{\mathrm P}$ with the same position's post-$m_0$ state increases masked-target CE by only $0.006$ when visible positions are modified and $0.004$ when masked positions are modified. Shuffling states within the same role and sequence is more damaging, but restoring each position's own post-$m_0$ state removes $95\%$ of the damage at masked positions and $98\%$ at visible positions. Thus, $s_{\mathrm P}$ retains useful information prepared early at each position while $s_{\mathrm E}$ continues to evolve.

\vspace{-4mm}

\paragraph{Masked targets gather visible-source influence during preservation.}
To test what the evolving stream carries during preservation, we mask a visible source $j$ and measure the loss increase at a masked target $i$. At a depth, we restore one or both streams at the source or target using states from the original run with the source visible, then rerun the remaining layers. Other positions retain their ablated-run states. Recovery is the fraction of the source-ablation loss increase removed by the patch. We analyze selected source--target pairs where removing the source degrades the target’s prediction.

Restoring the source tests whether later layers can use its restored state to repair the target prediction. Restoring the target tests whether useful source influence has already reached that position. Patching streams separately identifies which stream supports recovery. Full intervention and pair-selection details appear in Appendix~\ref{app:attn-intervention}.

\begin{figure}[H]
\centering
\includegraphics[width=0.8\linewidth]{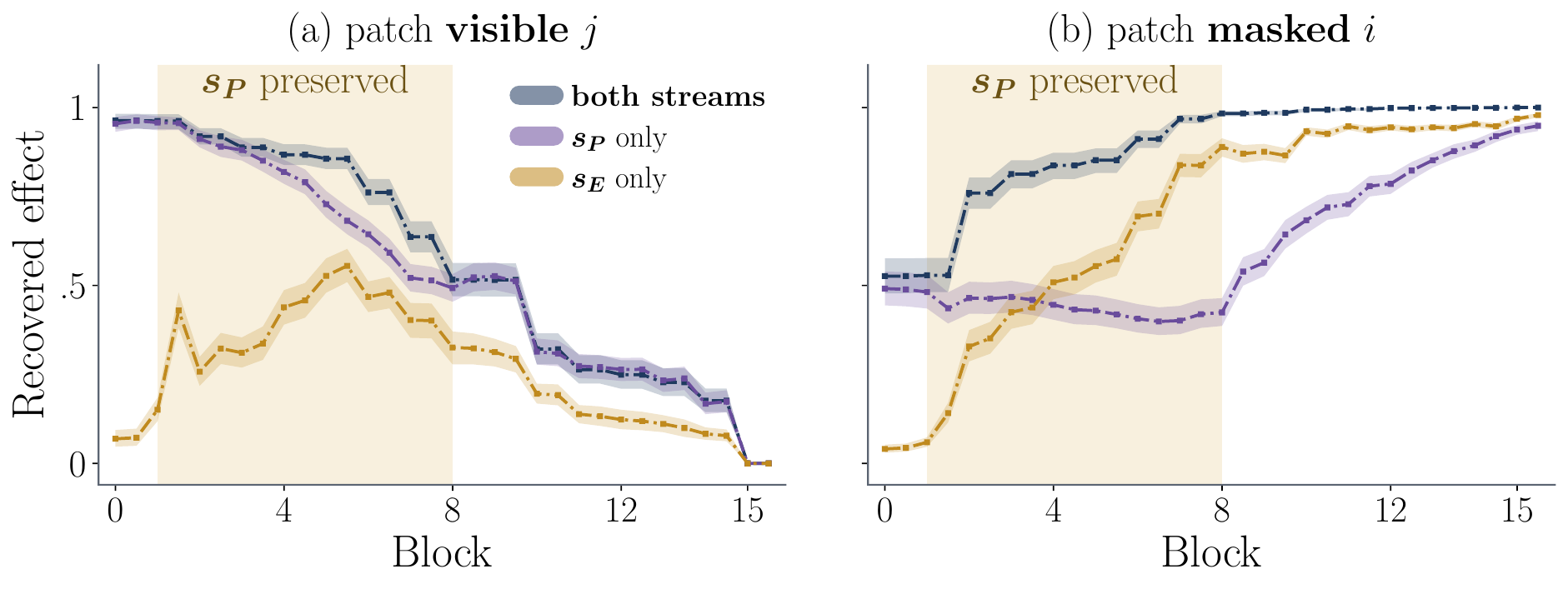}
\vspace{-1mm}
\caption{\textbf{Masked targets gather visible evidence in $s_{\mathrm E}$ during preservation.} Mean recovery falls when restoring both source streams (left) and rises when restoring target $s_{\mathrm E}$ (right). Bands show $95\%$ bootstrap intervals; shading marks $s_{\mathrm P}$ preservation. Details are in Appendix~\ref{app:attn-intervention}.}
\label{fig:evidence-transfer}
\end{figure}

\vspace{-1mm}

For selected pairs, visible-source influence is increasingly recoverable from the target's $s_{\mathrm E}$ during preservation. Restoring this stream reverses $5.9\%$ of the source-ablation effect at $a_1$, but $88.9\%$ at $a_8$ (Figure~\ref{fig:evidence-transfer}). By the interval's end, the target's evolving stream alone carries enough source-dependent information to recover most of the lost prediction benefit. Recovery from the target's $s_{\mathrm P}$ is substantial at the start but remains roughly flat, consistent with retaining its early context-dependent state.

Restoring both source streams recovers $96.2\%$ of the ablation effect at $a_1$ and $51.6\%$ at $a_8$: the network can use restored source evidence even after substantial information has accumulated at the target. Restoring the source's $s_{\mathrm P}$ alone yields substantial target recovery during preservation (Figure~\ref{fig:evidence-transfer}a).

\vspace{-4mm}

\paragraph{Late layers reuse and transform the preserved stream.}
Branches resume reading and writing $s_{\mathrm P}$ around $m_8$--$m_{10}$ (Figure~\ref{fig:circuit-atlas}). Its similarity to the post-$m_0$ state then declines, marking a transition from preservation to further transformation. We test late contributions by replacing both states at selected positions with the current $s_{\mathrm P}$ or $s_{\mathrm E}$. After $m_{12}$, visible-position duplication increases CE by $+0.010$ for $s_{\mathrm P}$ versus $+0.148$ for $s_{\mathrm E}$. Thus, late visible $s_{\mathrm P}$ supports downstream denoising with little loss, while masked positions benefit from both states. Together with renewed late reading, this links preservation to reuse of retained visible information and masked-prediction refinement using complementary stream states. Full results are in Appendix~\ref{app:state-preservation}.

\vspace{-4mm}

\paragraph{Preservation and renewed late reading recur across stream counts.}
Diffusion models with two to four streams share a pattern at masked and visible positions: one stream stays close to its first-block state, with little middle-depth reading and renewed late reading. Three- and four-stream AR show similar patterns, with more middle-depth reading and earlier departure from the early state. In two-stream AR, reading continues throughout the interval (Appendix~\ref{app:ar-streams}). Two-stream interventions also establish evidence gathering at masked targets during preservation.

\vspace{-4mm}

\paragraph{The functional stages recur in a 1B-parameter model.}
The separately trained 32-block 1B model repeats early preparation, position-specific preservation, source-to-target transfer, and late visible-state reuse (Appendix~\ref{app:scale-transfer}). Preparation extends beyond the first block, and the preserved stream is $s_0$. The same functional sequence therefore recurs with different stream identities and stage boundaries. At masked targets, useful source influence is available in both streams.

\begin{takeawaybox}
\textbf{Takeaway.} Denoising shows a preservation--transfer--reuse pattern: early blocks prepare position-specific states, one stream retains useful information as masked targets gather context, and late layers reuse retained visible states. Preservation and renewed late reading recur across stream counts and are stronger in diffusion than in matched AR controls.
\end{takeawaybox}


\enlargethispage{2\baselineskip}
\section{Conclusion}
\vspace{-1mm}
Our findings show that residual architecture shapes both model quality and how diffusion language models preserve evidence and refine predictions. Multiple persistent states retain useful early information while others evolve, and learned routing returns it to active computation later. This links visible and masked token roles to information processing through depth. We hope this work encourages closer study of residual architectures and pathways tailored to denoising, making when information is retained, updated, and reused an explicit consideration in diffusion-model design.



\brainprintbibliography

\begin{appendixpart}
\section{Experimental protocol and residual coordinates}
\label{app:protocol}

\paragraph{Pretraining and validation.} All width and FRR comparisons use a 16-block Qwen-style Transformer with 1,024 hidden dimensions, 2,816 MLP dimensions, 16 query heads, four key--value heads, and tied embeddings for the 152,064-token Dream vocabulary (tokenizer revision \texttt{ebb8da7f}). The ordered training sample is from FineWeb-Edu 10BT (revision \texttt{87f09149}). Each run uses 1,024-token sequences, 53,400 optimizer updates, and 65,536 effective tokens per update: 3,499,622,400 nominal training tokens. Each block contains one attention and one MLP residual interface, giving 32 routing sites. The mechanistic analyses use the final $n=2$ checkpoint at update 53,399.

\paragraph{Optimization and initialization.} The 337M runs use fused AdamW with $\beta=(0.9,0.95)$ and $\epsilon=10^{-8}$. The learning rate warms up linearly for 300 updates to $3\times10^{-4}$, then follows cosine decay to $3\times10^{-5}$. Global gradient-norm clipping uses a threshold of 1.0. Training and the permutation of training blocks use seed 1337. All learned-interface MDLM and AR runs ($n=1,2,3,4$), original 337M standard-residual baselines and FRR apply weight decay 0.1 to parameters with at least two dimensions and zero decay to the remaining parameters.

Linear and embedding weights are initialized from $\mathcal{N}(0,0.02^2)$, linear biases to zero, and RMSNorm scales to one. In the full router, read, write, and mixer head weights start at zero, and dynamic-head scale factors start at 0.01. Mixer biases are zero for the identity permutation and $-8$ for the other permutations. Read and write biases are $-1$ except for one stream with bias $+1$; this preferred stream alternates across interfaces. Stream embeddings are initialized to zero and frozen, so the initial stream states replicate the backbone input. FRR constructs the same initial backbone before replacing its interfaces with the factorization described in Appendix~\ref{app:frr}.

At training time, each sequence draws a corruption level $t\sim\mathrm{Uniform}(0,1)$, floored at 0.02. Tokens are masked independently with probability $t$, with one position forced masked if the draw produces none. Corruption levels and masks use seed $1{,}337{,}000{,}000$ plus the microbatch index. The MDLM objective averages the masked token CE sum divided by $1{,}024t$ across sequences. This is the same corruption and normalization rule for all MDLM interfaces. AR uses causal attention and next-token loss without corruption masks.

The in-domain validation comparison uses the same 64 held-out blocks and fixed masks for each model. Its MDLM score is the equally weighted mean of masked-reconstruction CE at mask probabilities $r\in\{0.1,0.3,0.5,0.7,0.9\}$. At each ratio, each sequence's summed masked-target loss is divided by the expected masked count $1{,}024r$, then averaged over the 64 sequences. Reconstruction perplexity denotes the exponential of the corresponding reconstruction score. During fixed-corruption evaluation, the training corruption parameter is set to $t=r$. AR validation uses the $64\times1{,}023$ next-token targets. Within each objective, the same validation protocol is applied at the same training step.

\paragraph{External-corpus evaluation.} We use \path{Skylion007/openwebtext} (\texttt{plain\_text}, train), \path{Salesforce/wikitext} (\texttt{wikitext-103-raw-v1}, test), \path{EleutherAI/lambada_openai} (\texttt{en}, test), and \path{allenai/c4} (\texttt{en}, validation). Dataset revisions are not pinned; the Dream tokenizer revision is \texttt{ebb8da7f}. Records retain their dataset order without shuffling. OpenWebText and C4 are streamed from their first up to 2,000 and 3,000 records, respectively; WikiText-103 and LAMBADA follow split order. Empty or whitespace-only texts are skipped. Each text is tokenized without additional special tokens and followed by \texttt{eos\_token\_id}. The concatenated stream is cut into 256 full 1,024-token blocks per corpus, stopping once these blocks are obtained; a final incomplete block is excluded. All models receive the same blocks and masks at $r\in\{0.1,0.5,0.9\}$, with mask seeds 901, 905, and 909. We calculate actual-masked-token CE separately at each ratio and average the three CE values. The relative improvement shown in the main text is $100[1-\exp(\ce_{\rm model}-\ce_{\rm residual})]$. Paired intervals use 10,000 sequence resamples, shared across models and ratios. All four corpora use this masked-reconstruction protocol, including LAMBADA.

\paragraph{Causal interventions.} Unless stated otherwise, calibration means are computed in float64 on 64 sequences disjoint from evaluation, separately at each attention and MLP interface. The 337M model has 32 interfaces, and the 1B model has 64. The 337M role and volume interventions use 1,024 common held-out sequences at $r=0.5$; the corresponding 1B setup uses 128 held-out sequences. Within each comparison, every condition uses the same corrupted inputs and masked targets. Role labels are read from the corrupted input, never from the clean target. Interventions target specified router coordinates at selected tokens and interfaces; unselected controls are computed normally from the current states, and subsequent states and router outputs are recomputed normally. Read interventions replace the pre-mapping $\mathcal{H}_{\ell}^{\mathrm{pre}}$; read shares in the figures express its relative weights across streams. In a role swap, masked positions receive the visible calibration mean and vice versa. The instrumented intact path was checked against the original logits in FP32 and BF16.

The causal splits use fixed block IDs. At 337M, calibration uses blocks 128--191, discovery uses 256--383, and confirmation uses 1024--2047. At 1B, calibration uses 128--191 and evaluation uses 1024--1151. Corruption masks use the block-specific seed $782+1009\,\mathrm{block\_id}+10017\,\mathrm{round}(100r)$. Volume-shuffle runs use seeds 130921 and 130922, with permutations formed separately by block, interface, and role.

For sequence $s$, let $L_s(c)$ be its summed target loss under condition $c$, and $N_s$ its valid target count. Each of the 5,000 paired sequence-bootstrap resamples $B$ computes
\begin{equation} \widehat{\ce}_B(c)=\frac{\sum_{s\in B}L_s(c)}{\sum_{s\in B}N_s},\qquad \Delta_B(c)=\widehat{\ce}_B(c)-\widehat{\ce}_B(\mathrm{intact}). \label{eq:app-estimator} \end{equation}
Intervals and gain ratios are calculated within the same paired resample.

\paragraph{Routing-profile aggregation.} For each interface and token role, read share is $\overline{h}_i/\sum_j\overline{h}_j$, where $\overline{h}_i$ averages that stream's pre-mapping weight across all positions of the role. Positions receive equal weight, so each sequence contributes according to its number of positions in that role. The main 337M routing profile uses the 64 calibration sequences, containing 32,529 masked and 33,007 visible positions.

\paragraph{Write coordinates and volume interventions.} The routing interventions target the full mHC-lite checkpoints, whose physical write gains lie in $[0,2]$. For two-stream write interventions, let $V=g_0+g_1=2v_\ell$ and, for $V>0$, $q_i=g_i/V$. These coordinates are computed from the current router output in the intervened forward pass. Calibration means are computed separately for volume and normalized destination; destination means average $q_i=g_i/V$ directly, rather than taking the ratio of mean gains to mean volume.

A proposed volume $V'$ is applied through $g_i'=V'q_i=(V'/V)g_i$, preserving normalized destination. The replacement is accepted only when $V>0$, $V'>0$, and the proposed gains are finite and lie in $[0,2]$. For finite positive volumes and normalized destination, the upper-bound condition is $V'\max_i q_i\leq2$. Otherwise, both gains remain unchanged. Accepted replacements therefore preserve $q$ and keep gains in $[0,2]$ without clipping. Mean-based volume interventions propose the calibrated global or role-conditioned mean for the interface, using the opposite role's mean for role swaps. Role-mean replacement removes within-role volume variation at positions where the proposal is accepted.

Within-role shuffling uses random disjoint pairs of positions, formed separately within each sequence, role, and interface. For a pair $a,b$, we propose $g_{a,i}'=V_bq_{a,i}$ and $g_{b,i}'=V_aq_{b,i}$. The exchange is accepted only when both original volumes are positive and both proposed gain pairs are finite and satisfy the gain bounds: $V_b\max_i q_{a,i}\leq2$ and $V_a\max_i q_{b,i}\leq2$. Rejected pairs and unpaired positions remain unchanged. Thus, shuffling preserves each token's normalized destination and the complete multiset of volumes within each group, to numerical precision.

\paragraph{Destination and joint interventions.} Destination interventions replace normalized destination $q$ while retaining the token's current volume $V$. Joint interventions replace all four controls---read, mixing, normalized write destination, and write volume---with their prescribed calibration means. The proposed write gains are reconstructed as $g_i'=V'q_i'$, with any unselected write coordinate retained from the current router output. The proposed gain pair is applied only when both reconstructed gains are finite and lie in $[0,2]$; otherwise, both current gains remain unchanged. No clipping is used. In joint interventions, rejection of the write-gain pair does not undo the read and mixing replacements.

For corruption-level sweeps, role means are recalibrated on disjoint calibration sequences at each mask ratio and remain fixed during evaluation. Numerical checks verify normalized-destination preservation and gain bounds for volume interventions, as well as volume-multiset preservation under shuffling.

\paragraph{Volume-intervention coverage.} At each interface, acceptance coverage is the fraction of selected valid token positions for which the proposed volume replacement satisfies the acceptance rule. Accepted proposals are counted even when the proposed volume equals the original value. For within-role shuffling, the numerator includes both positions of each mutually feasible pair; rejected pairs and unpaired positions remain in the denominator. Padding is excluded. Table~\ref{tab:volume-coverage} reports coverage averaged across interfaces at $r=0.5$.

\begin{table}[H]
\centering
\small
\caption{Mean interface-wise acceptance coverage (\%) for volume interventions at $r=0.5$. Global mean and role swap target both roles; the other rows target the specified role.}
\label{tab:volume-coverage}
\begin{tabular}{@{}lrr@{}}
\toprule
Intervention & 337M & 1B \\
\midrule
Role mean, masked & 97.48 & 99.51 \\
Role mean, visible & 98.22 & 99.58 \\
Within-role shuffle, masked & 93.33 & 97.09 \\
Within-role shuffle, visible & 93.15 & 97.04 \\
Global mean, both roles & 97.59 & 99.48 \\
Role swap, both roles & 95.15 & 99.10 \\
\bottomrule
\end{tabular}
\end{table}

Mean coverage exceeds 93\% for role-mean replacement and within-role shuffling at both scales. Volume role-swap coverage is 1.38\% at one masked-position interface in the 337M model. CE effects are measured for the accepted-replacement procedures defined above.

\paragraph{Destination and joint acceptance coverage.} We measured acceptance during full-network intervention replays at $r=0.5$, targeting both token roles at every attention and MLP interface. A position is accepted when both proposed gains are finite and lie in $[0,2]$; numerically unchanged accepted proposals are counted. Rejected proposals retain both current gains. In joint interventions, rejection of the gain pair does not undo the read and mixing replacements. Coverage is the mean of interface-wise accepted/selected percentages ($K=32$ for 337M; $K=64$ for 1B). For the ``All'' column, role counts are pooled within each interface before averaging. Padding is excluded. The replays reproduced the corresponding CE estimates reported in Table~\ref{tab:role-full} and Appendix~\ref{app:scale-allocation}.

\begin{table}[H]
\centering
\small
\caption{Acceptance coverage (\%) for destination and joint interventions using normalized write destination $q_i=g_i/(g_0+g_1)$.}
\label{tab:destination-joint-coverage}
\begin{tabular}{@{}llrrr@{}}
\toprule
Model & Intervention & Masked & Visible & All \\
\midrule
337M & Destination, global mean & 93.84 & 98.45 & 96.15 \\
337M & Destination, role means & 97.00 & 97.61 & 97.31 \\
337M & Destination, role swap & 91.04 & 98.48 & 94.76 \\
337M & Joint, global mean & 100.00 & 100.00 & 100.00 \\
337M & Joint, role means & 100.00 & 100.00 & 100.00 \\
337M & Joint, role swap & 100.00 & 100.00 & 100.00 \\
\midrule
1B & Destination, global mean & 99.30 & 99.47 & 99.39 \\
1B & Destination, role means & 98.93 & 99.44 & 99.18 \\
\bottomrule
\end{tabular}
\end{table}

Joint acceptance was 100.00\% in the measured runs.
\section{Stream width and transfer}
\label{app:width}

\begin{table}[H]
\centering
\small
\caption{Final in-domain validation at update 53,399. Learned $n=1$
retains token-adaptive scalar gates. Lower CE is better.}
\label{tab:width-full}
\begin{tabular}{@{}lcrr@{}}
\toprule
Interface & Parameters & MDLM CE & AR CE\\
\midrule
Standard residual & 336.1M & 3.605514 & 3.018551\\
Learned $n=1$ & 336.2M & 3.580209 & 3.014518\\
$n=2$ & 336.6M & 3.546667 & 2.986202\\
$n=3$ & 337.4M & \textbf{3.541492} & 2.966527\\
$n=4$ & 340.4M & 3.542203 & \textbf{2.957888}\\
\bottomrule
\end{tabular}
\end{table}

At $n=1$, the residual mixer reduces to the identity; the learned pre- and post-mappings are token-dependent scalar read and write gates.

Relative to learned $n=1$, adding a second stream reduces validation CE by $0.033542$, capturing $86.6\%$ of the largest additional reduction ($0.038717$ at $n=3$). Relative to the standard-residual baseline, $n=2$ reduces CE by $0.058847$, or $91.9\%$ of the largest observed reduction ($0.064022$ at $n=3$). The best wider model improves on $n=2$ by a further $0.005175$ CE. These percentages use different reference models; the main text reports the gain beyond learned $n=1$ to quantify the benefit of additional streams.

\begin{table}[H]
\centering
\small
\caption{External masked-reconstruction CE change from the standard
residual, with paired 95\% sequence-bootstrap intervals. All $n\geq2$
models improve on every corpus; $n=2$ has the largest mean percentage
reduction across corpora.}
\label{tab:width-transfer}
\begin{tabular}{@{}llrr@{}}
\toprule
Corpus & Streams & $\Delta$CE & 95\% interval\\
\midrule
OpenWebText & $n=1$ & $-.0145$ & [$-.0190,-.0100$]\\
 & $n=2$ & $-.0472$ & [$-.0521,-.0423$]\\
 & $n=3$ & $-.0566$ & [$-.0618,-.0516$]\\
 & $n=4$ & $-.0528$ & [$-.0583,-.0476$]\\
\midrule
WikiText-103 & $n=1$ & $-.0118$ & [$-.0178,-.0058$]\\
 & $n=2$ & $-.0791$ & [$-.0848,-.0735$]\\
 & $n=3$ & $-.0635$ & [$-.0702,-.0564$]\\
 & $n=4$ & $-.0336$ & [$-.0395,-.0275$]\\
\midrule
LAMBADA & $n=1$ & $+.0015$ & [$-.0026,+.0057$]\\
 & $n=2$ & $-.0306$ & [$-.0350,-.0263$]\\
 & $n=3$ & $-.0230$ & [$-.0275,-.0184$]\\
 & $n=4$ & $-.0271$ & [$-.0313,-.0229$]\\
\midrule
C4 & $n=1$ & $-.0178$ & [$-.0224,-.0133$]\\
 & $n=2$ & $-.0544$ & [$-.0600,-.0488$]\\
 & $n=3$ & $-.0531$ & [$-.0584,-.0480$]\\
 & $n=4$ & $-.0554$ & [$-.0611,-.0499$]\\
\bottomrule
\end{tabular}
\end{table}

The $n=2$ improvement is positive on all four external corpora. Its
relative reconstruction-perplexity reductions are 4.61\% on OpenWebText, 7.61\%
on WikiText-103, 3.02\% on LAMBADA, and 5.29\% on C4, for a 5.13\%
unweighted mean. Larger width improves some individual corpora, while
$n=2$ gives the strongest mean transfer in this sweep.

\section{Role-factorized routing learned from scratch}
\label{app:frr}

FRR trains two persistent streams with a compact interface. At each attention and MLP site, each role $\rho\in\{\mathrm{masked},\mathrm{visible}\}$ has free read logits $\boldsymbol{u}_{\ell,\rho}$ and destination logits $\boldsymbol{z}_{\ell,\rho}$. Read weights are $h_{\ell,\rho,i}=\sigma(u_{\ell,\rho,i})$, without sum normalization; destination is $\boldsymbol{d}_{\ell,\rho}=\mathrm{softmax}(\boldsymbol{z}_{\ell,\rho})$. The masked and visible read logits start equal, copied from the initial full-router read bias.

The token-dependent volume predictor takes the concatenated current states $[s_0;s_1]\in\mathbb{R}^{2048}$:
\begin{equation}
v_\ell=2\sigma\!\left(\alpha_\ell W_\ell\,\mathrm{RMSNorm}([s_0;s_1])+b_\ell\right).
\end{equation}
RMSNorm uses $\epsilon=10^{-6}$, followed by one linear projection to a scalar and the sigmoid. The predictor starts with $W_\ell=0$, $b_\ell=0$, and $\alpha_\ell=0.01$, giving initial volume $v_\ell=1$. Physical gains are $g_i=2v_\ell d_i$, and the old-state mixer is the identity. Thus $0<v_\ell<2$, $0<g_i<4$, and $\sum_i g_i=2v_\ell<4$. FRR has a per-stream gain bound of 4; the full mHC-lite router and its intervention protocol use a bound of 2. Both parameterizations retain the same write-volume range $0<v_\ell<2$.

FRR tests whether role-conditioned routing together with dynamic volume can guide the learning of useful two-stream routes. It is trained from scratch with the same training seed, data order, corruption masks, backbone, learning-rate schedule, and validation blocks as the width comparison. Its optimizer parameter groups match full $n=2$, and backbone and router parameters share the scheduled learning rate with peak $3\times10^{-4}$.

\begin{table}[H]
\centering
\small
\caption{Final MDLM validation CE by corruption ratio. The mean weights
the five ratios equally.}
\label{tab:frr}
\begin{tabular}{@{}lrrrrrr@{}}
\toprule
Interface & $r=.1$ & $r=.3$ & $r=.5$ & $r=.7$ & $r=.9$ & Mean\\
\midrule
Residual & 1.7060 & 2.2986 & 3.2581 & 4.5212 & 6.2436 & 3.605514\\
Learned $n=1$ & 1.6760 & 2.2593 & 3.2229 & 4.4990 & 6.2439 & 3.580209\\
FRR & 1.6518 & 2.2550 & 3.2054 & 4.4759 & 6.2288 & 3.563375\\
Full $n=2$ & 1.6212 & 2.2361 & 3.1823 & 4.4649 & 6.2289 & 3.546667\\
\bottomrule
\end{tabular}
\end{table}

FRR gains $0.042139$ CE over the standard residual, reproducing
71.6\% of the full $n=2$ gain. It also improves on learned $n=1$
by $0.016834$ CE, closing 50.2\% of the gap from $n=1$ to full
$n=2$. Its advantage over the standard residual is present at every
recorded validation milestone from 5,000 updates onward. The trained
comparison supports the joint usefulness of a second stream, role-based
read and write destination, and a state-dependent scalar volume. Full $n=2$ provides a further $0.016708$ CE improvement.

\section{Role routing in the full two-stream checkpoint}
\label{app:role-table}

At $r=0.5$, intact $n=2$ has CE 3.20422 and the separately trained
standard residual has CE 3.26993 on the same 1,024 test sequences. In Table~\ref{tab:role-full}, each condition targets one control throughout the network while the others remain dynamic; write-gain proposals are applied only where the reconstruction satisfies the gain bounds. The joint row changes the four controls together. Role rescue is the fraction
$(\ce_{\rm global}-\ce_{\rm role})/
(\ce_{\rm global}-\ce_{\rm intact})$.

\begin{table}[H]
\centering
\small
\caption{Increase in masked-target CE over intact $n=2$, with paired
95\% intervals. Global means discard role; role means retain
masked/visible assignment; swaps exchange it.}
\label{tab:role-full}
\begin{tabular}{@{}llrr@{}}
\toprule
Control & Setting & $\Delta$CE & 95\% interval\\
\midrule
Read & Global & $+.9493$ & [$+.9406,+.9575$]\\
 & Role & $+.0405$ & [$+.0392,+.0419$]\\
 & Swap & $+1.8108$ & [$+1.7970,+1.8241$]\\
\midrule
Write destination & Global & $+1.2551$ & [$+1.2391,+1.2712$]\\
 & Role & $+.0171$ & [$+.0162,+.0181$]\\
 & Swap & $+2.9771$ & [$+2.9471,+3.0062$]\\
\midrule
Write volume & Global & $+2.0262$ & [$+2.0044,+2.0471$]\\
 & Role & $+.2003$ & [$+.1959,+.2046$]\\
 & Swap & $+3.1725$ & [$+3.1431,+3.2008$]\\
\midrule
Mixing & Global & $+.0194$ & [$+.0183,+.0204$]\\
 & Role & $+.0058$ & [$+.0053,+.0063$]\\
 & Swap & $+.0431$ & [$+.0415,+.0446$]\\
\midrule
All jointly & Global & $+4.5115$ & [$+4.4759,+4.5465$]\\
 & Role & $+1.6453$ & [$+1.6242,+1.6659$]\\
 & Swap & $+4.7515$ & [$+4.7180,+4.7848$]\\
\bottomrule
\end{tabular}
\end{table}

Read and destination show the clearest role structure: role means
recover 95.7\% and 98.6\% of the corresponding global-mean damage,
respectively. Swapping the assignments makes both substantially more
damaging. Write volume additionally carries a consequential
$+0.2003$ CE within-role component. Joint replacement targets all four controls simultaneously with their prescribed calibration means. Joint write-gain acceptance was 100.00\% in these runs (Table~\ref{tab:destination-joint-coverage}). The role-mean CE increase of $1.6453$ exceeds any of the individual role-mean effects.

\section{Token-matched write volume and depth}
\label{app:volume}

\paragraph{Matching volume to tokens.} A role-mean replacement removes within-role volume variation at positions where the proportional replacement satisfies the gain bounds. Within-role shuffling instead exchanges volumes between random disjoint pairs within each sequence, role, and interface, accepting an exchange only when both directions are feasible. Rejected pairs remain unchanged, preserving the complete volume multiset within each group. Both interventions retain each affected token's current normalized write destination, $q_i=g_i/V$, following Appendix~\ref{app:protocol}. Read and mixing remain dynamic; subsequent states and controls are recomputed normally. Table~\ref{tab:volume-assignment} shows that token assignment matters beyond the role-specific volume distribution. Acceptance coverage for both scales is reported in Table~\ref{tab:volume-coverage}.

\begin{table}[H] \centering \small \caption{Volume interventions at $r=0.5$: paired CE increases over the intact checkpoint, with 95\% sequence-bootstrap intervals on 1,024 held-out sequences.} \label{tab:volume-assignment} \begin{tabular}{@{}lrr@{}} \toprule Positions changed & Role mean & Within-role shuffle \\ \midrule Masked & $+.0684$ [$+.0665,+.0703$] & $+.1322$ [$+.1300,+.1345$] \\ Visible & $+.1242$ [$+.1206,+.1276$] & $+.2939$ [$+.2886,+.2992$] \\ \bottomrule \end{tabular} \end{table}

Shuffling preserves the current volume multiset locally at each intervened interface; subsequent router outputs are recomputed from the intervened states. Shuffling tests token assignment at fixed local volume values; averaging tests replacement by a role-conditioned constant.

\paragraph{Localization through depth.}
Replacing volume only in the first block raises CE by $+.0673$
[$+.0651,+.0694$] at visible positions and $+.0015$
[$+.0013,+.0018$] at masked positions. In the last block the pattern
reverses: masked replacement raises CE by $+.0181$
[$+.0172,+.0190$], while visible replacement has no measured effect. The final-block visible effect is structurally zero because no subsequent attention layer can transmit these changes to masked targets. Across interior blocks, visible-token sensitivity declines sharply after the first block, while masked-token sensitivity follows a non-monotonic profile that peaks near the readout.
The complete per-block intervention is plotted in
Figure~\ref{fig:volume-schedule}b. The profile links early
visible-evidence processing with later masked-target refinement.

\paragraph{Single-block acceptance coverage.} Table~\ref{tab:volume-depth-coverage} reports acceptance coverage for the same single-block role-mean volume interventions used in Figure~\ref{fig:volume-schedule}b. Coverage is averaged only over the attention and MLP interfaces of the selected block. Masked-only and visible-only columns correspond to separate intervention runs.

\begin{table}[H]
\centering
\small
\caption{Mean acceptance coverage (\%) for single-block volume interventions in the 337M model at $r=0.5$. Blocks are indexed from zero; the first row applies to each block individually.}
\label{tab:volume-depth-coverage}
\begin{tabular}{@{}lrr@{}}
\toprule
Block & Masked-only & Visible-only \\
\midrule
0--9 & 100.00 & 100.00 \\
10 & 100.00 & 98.79 \\
11 & 100.00 & 99.19 \\
12 & 100.00 & 100.00 \\
13 & 95.13 & 99.37 \\
14 & 86.29 & 100.00 \\
15 & 81.08 & 79.92 \\
\bottomrule
\end{tabular}
\end{table}

\begin{table}[H]
\centering
\small
\caption{Network-wide role-mean volume replacement at different mask
ratios. Each entry is a paired CE increase over its intact checkpoint;
128 sequences are used per ratio.}
\label{tab:volume-corruption}
\begin{tabular}{@{}rrrr@{}}
\toprule
Mask ratio & Masked positions & Visible positions & Both roles\\
\midrule
0.1 & $+.0481$ & $+.1220$ & $+.1769$\\
0.3 & $+.0556$ & $+.1376$ & $+.2014$\\
0.5 & $+.0621$ & $+.1256$ & $+.1972$\\
0.7 & $+.0547$ & $+.0882$ & $+.1501$\\
0.9 & $+.0302$ & $+.0349$ & $+.0672$\\
\bottomrule
\end{tabular}
\end{table}

\paragraph{Acceptance across corruption levels.} Table~\ref{tab:volume-sweep-coverage} reports acceptance coverage for the corresponding corruption-sweep interventions in Table~\ref{tab:volume-corruption}. Each column is measured in its own intervention run, with coverage averaged across the 32 interfaces. Both-role coverage is measured directly, rather than obtained by combining the two single-role conditions. Mean acceptance remains above 97\% in every condition.

\begin{table}[H]
\centering
\small
\caption{Mean acceptance coverage (\%) for network-wide role-mean volume replacement in the 337M model across mask ratios.}
\label{tab:volume-sweep-coverage}
\begin{tabular}{@{}rrrr@{}}
\toprule
Mask ratio & Masked-only & Visible-only & Both roles \\
\midrule
0.1 & 97.54 & 98.50 & 98.45 \\
0.3 & 97.45 & 98.27 & 98.16 \\
0.5 & 97.36 & 98.20 & 97.99 \\
0.7 & 97.28 & 98.03 & 97.77 \\
0.9 & 97.45 & 98.38 & 97.79 \\
\bottomrule
\end{tabular}
\end{table}

The full-network effect peaks under partial observation and declines
toward $r=0.9$. The both-roles intervention is measured directly;
it is not obtained by adding the two single-role effects.

\subsection{Autoregressive control}
\label{app:ar}

State-dependent volume also affects AR checkpoints. On 1,024 matched
AR sequences, replacing volume by an interface mean raises CE by
$+.5837$ [$+.5756,+.5918$] for $n=2$, $+.3648$
[$+.3552,+.3745$] for $n=3$, and $+.3831$
[$+.3745,+.3918$] for $n=4$. Even exact-position means leave
effects of $+.5466$, $+.3574$, and $+.3631$, respectively. This
control identifies dynamic volume as a general residual mechanism;
the distinctive diffusion structure is the strong causal assignment
of read and write routes by observed corruption role.

\section{First-block construction and evidence transfer}
\label{app:circuit}
\label{app:attn-intervention}

\paragraph{Addressing the first update.}
Deletion of the $a_0$ write, restoration of its contribution to the $m_0$ input, and write rerouting use the same 1,024 sequences (blocks 1024--2047), containing 523,845 masked targets, and the same corruption masks, with matching saved mask SHA-256 fingerprints.
The first attention interface, $a_0$, sends most of its write to
$s_{\mathrm E}$ at visible tokens (write shares computed from role-mean gains $(0.88,0.12)$) and to $s_{\mathrm P}$ at masked tokens
($(0.02,0.98)$). In the masked-role edge intervention, deleting
the $a_0$ write raises CE from 3.2043 to 8.0761. Restoring only
its contribution to the first MLP input, $m_0$, gives CE 3.2900,
recovering 98.24\% of the deletion effect. A separate destination
test keeps the branch output and total gain fixed while reassigning
the two write gains. Sending the visible $a_0$ write to the opposite
stream costs $+3.0416$ CE; sending it to the learned stream costs
only $+.0271$. The corresponding masked-role effects are $+.4623$
and $+.0017$. These tests establish that the early state depends
causally on where the update is placed.

\paragraph{Source-to-target patching.}
The search uses the 128 discovery sequences. We first evaluate all 65,122 masked positions. In each sequence, we retain correctly predicted targets and select the six with the highest correct-token log probability, giving 768 targets for source-influence testing. For each selected target, we mask candidate visible sources and retain pairs for which removing the source lowers the target's correct-token log probability by at least $0.5$ nats. The candidate pool contains six nearest visible positions and two random positions; we select the source with the largest effect. This gives $179$ pairs across $91$ sequences.

For each pair, we compare the original run, where the source is visible, with an ablated run, where it is masked. After each interface, we replace $s_{\mathrm E}$, $s_{\mathrm P}$, or both at either the source or target with their states from the original run, then rerun the remaining network. All unpatched states at that interface retain their ablated-run values. Source patching tests whether the remaining layers can use restored source evidence; target patching tests whether useful source influence is already available in the target state.

Let $\ell_{ij}^{\rm clean}$, $\ell_{ij}^{\rm ablated}$, and $\ell_{ij}^{\rm patch}$ denote the correct-token loss at target $i$ in the original, ablated, and patched runs for source $j$. Pairwise recovery is
\begin{equation}
 R_{ij}=\frac{\ell_{ij}^{\rm ablated}-\ell_{ij}^{\rm patch}}{\ell_{ij}^{\rm ablated}-\ell_{ij}^{\rm clean}}.
 \label{eq:app-patch-recovery}
\end{equation}
Figure~\ref{fig:evidence-transfer} reports mean recovery over the selected pairs. We obtain 95\% confidence intervals from 5,000 sequence-cluster bootstrap resamples, sampling the 91 sequences with replacement and retaining all pairs from each sampled sequence, including multiplicities. Recovery is averaged over the resulting pairs, and the same resamples are used across depths and patch conditions. Recovery averages the pairwise ratios $R_{ij}$; routing interventions use the target-count-weighted CE estimator in Equation~\ref{eq:app-estimator}.
During preservation, recovery from restoring the target's $s_{\mathrm E}$ rises from $0.059$ at $a_1$ to $0.889$ at $a_8$. By the end of this interval, its evolving state carries enough source-dependent information to repair most of the source-ablation effect. Recovery from the target's $s_{\mathrm P}$ is already substantial but stays near $0.4$--$0.5$ until the later reopening phase, consistent with retaining an early context-dependent state. Restoring both target streams at $a_8$ recovers $0.983$.

Recovery from restoring both source streams falls from $0.962$ at $a_1$ to $0.516$ at $a_8$. Substantial recovery remains possible at this later depth, showing that the remaining network can still use restored source evidence. Together, these interventions locate useful visible-source influence in the target's evolving stream during preservation, while showing that later computation retains access to source evidence.

\section{Preservation and late reuse of the second stream}
\label{app:state-preservation}

\begin{figure}[H]
\centering
\includegraphics[width=0.96\linewidth]{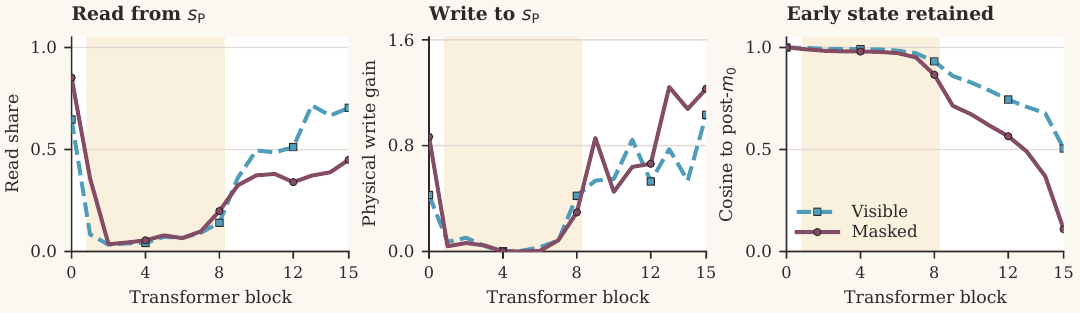}
\caption{\textbf{A first-block state persists through the middle network.}
The preserved stream receives little read or write mass through middle
depth, while its cosine similarity to its own post-$m_0$ state remains
high at masked and visible positions. Later read and write activity
coincides with transformation of this sample-specific state.}
\label{fig:preserved-profile}
\end{figure}

Own-state replacements and within-role state shuffles use 128 sequences (blocks 256--383). State shuffling uses a deterministic cyclic shift among the selected same-role positions within each sequence.
At $a_8$, $s_{\mathrm P}$ has cosine similarity 0.949 to its own
post-$m_0$ masked state and 0.965 at visible positions. Replacing
$s_{\mathrm P}^{a_8}$ with the \emph{same position's}
$s_{\mathrm P}^{m_0}$ raises CE by only $+.0037$
[$+.0029,+.0045$] for masked positions and $+.0059$
[$+.0048,+.0070$] for visible positions. Replacing it with the
post-$a_0$ state is much more disruptive, particularly at visible
positions ($+.5870$ CE), identifying the post-$m_0$ state as the
preserved object.

Position identity remains essential. Shuffling $s_{\mathrm P}$ among
same-role tokens within each sequence after $a_8$ raises CE by
$+.0820$ [$+.0777,+.0864$] for masked positions and
$+.3721$ [$+.3495,+.3962$] for visible positions. Restoring the
matched post-$m_0$ state removes 95\% and 98\% of this shuffle
damage, respectively. Thus, the state remains position-specific within the sequence.

Read and write mass on $s_{\mathrm P}$ increase again after roughly
$m_8$--$m_{10}$. Its masked-position cosine similarity to the
post-$m_0$ state falls from 0.949 at $a_8$ to 0.672 at $m_{10}$
and 0.109 at $m_{15}$, consistent with transformation during
reuse. Late role-specific interventions in the main text show
that visible positions can rely primarily on the transformed $s_{\mathrm P}$ for late continuation under stream duplication, while the two stream states are not interchangeable at masked positions.

\begin{table}[H]
\centering
\small
\caption{Late stream-content intervention after $m_{12}$. Both streams
at positions of the selected role are set to either their current $s_{\mathrm E}$ or their current $s_{\mathrm P}$ before the remaining network is rerun. Values are
paired CE increases over intact $n=2$, with 95\% intervals on the same 128 sequences (blocks 256--383) and corruption masks as the preservation tests.}
\label{tab:late-reuse}
\begin{tabular}{@{}lrr@{}}
\toprule
Role & Both set to $s_{\mathrm E}$ & Both set to $s_{\mathrm P}$\\
\midrule
Visible & $+.1475$ [$+.1356,+.1604$] & $+.0104$ [$+.0087,+.0121$]\\
Masked & $+.0772$ [$+.0741,+.0804$] & $+.2304$ [$+.2184,+.2432$]\\
\bottomrule
\end{tabular}
\end{table}

At visible positions, the transformed $s_{\mathrm P}$ is the stronger late carrier. At masked positions, either stream duplication degrades predictions, demonstrating functional non-interchangeability. Removing $s_{\mathrm P}$ immediately before the output
head adds $+.3549$ CE [$+.3402,+.3704$] on masked targets,
confirming its contribution at readout.

\section{Preservation across stream counts and the AR comparison}
\label{app:ar-streams}

We repeat the routing survey on 64 common sequences for diffusion
$n=2,3,4$ and their AR counterparts. For each model and role,
Table~\ref{tab:cross-width-preservation} identifies the stream with
the smallest mean read share in the $a_2$--$a_8$ window, then
measures its later read share over $a_{11}$--$m_{15}$.
The stream is selected by the middle-window rule before comparing
the late values.

\begin{table}[H]
\centering
\small
\caption{Read share of the candidate stream selected by minimum middle-depth reading.
Diffusion candidates have particularly low middle-depth read share
and are read more heavily near output.}
\label{tab:cross-width-preservation}
\begin{tabular}{@{}llrr@{}}
\toprule
Objective & Streams and role & Middle read share & Late read share\\
\midrule
MDLM & $n=2$, masked / visible & .062 / .059 & .386 / .617\\
MDLM & $n=3$, masked / visible & .062 / .086 & .598 / .639\\
MDLM & $n=4$, masked / visible & .051 / .075 & .301 / .368\\
AR & $n=2$ & .473 & .793\\
AR & $n=3$ & .167 & .632\\
AR & $n=4$ & .117 & .560\\
\bottomrule
\end{tabular}
\end{table}

\begin{figure}[H]
\centering
\includegraphics[width=0.88\linewidth]{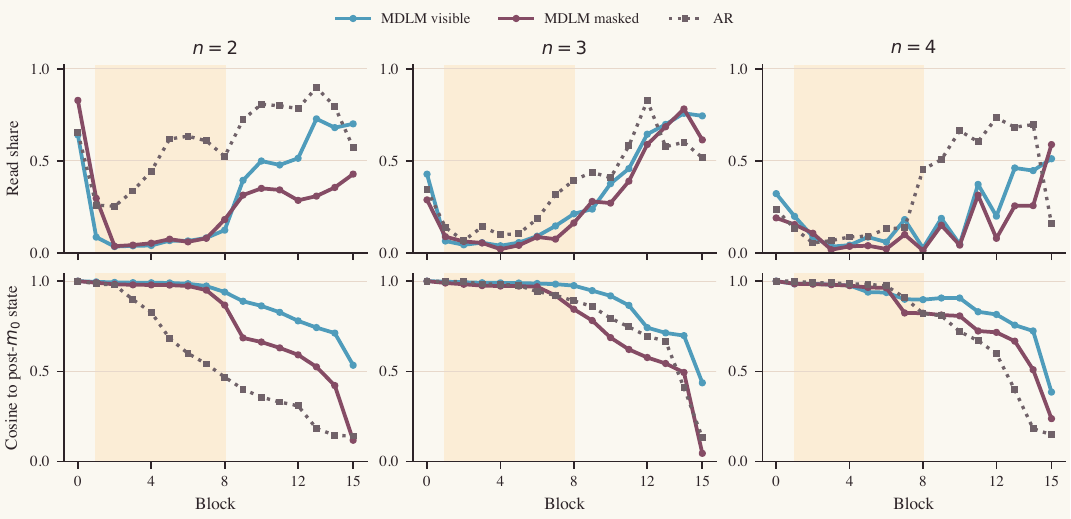}
\caption{\textbf{Diffusion shows stronger early-state preservation and lower middle-depth reading.} For each residual stream count, the top row gives the candidate stream's
read share and the bottom row its cosine similarity to the post-$m_0$
state. Diffusion is separated by visible and masked roles; AR positions are not
partitioned by a corruption mask. Shading marks the middle-depth preservation
interval.}
\label{fig:ar-preservation}
\end{figure}

The diffusion pattern recurs with two, three, and four residual streams at both masked and visible positions: the selected stream's middle-depth read share stays below $0.10$, then increases in late layers. The AR $n=2$ candidate is read throughout the middle network and departs rapidly from its first-block state. A related pattern of preservation followed by renewed late reading also appears in AR models with three and four streams, but their candidates are read more during the middle interval and depart sooner from the post-$m_0$ state (Figure~\ref{fig:ar-preservation}). The diffusion models studied therefore show lower middle-depth reading and more persistent early-state similarity than the matched AR controls, followed by renewed late reading. The evidence-transfer component is established separately by the source-to-target interventions on the two-stream diffusion model (Appendix~\ref{app:attn-intervention}).

\section{The 337M functional organization recurs at 1B}
\label{app:scale-transfer}

The 1B model repeats the 337M functional sequence: prepare position-specific
information early, retain it through middle depth while source influence
reaches masked targets, and reuse retained visible states late. We align
streams by their preservation and late-reuse functions: $s_P$ is $s_1$ at
337M and $s_0$ at 1B.

\paragraph{Setup.}
The 1B model (1.028B parameters, 32 blocks) was trained for 124,000 steps
on 8.127B input tokens. Its hidden and MLP dimensions are 1,536 and 4,096, with 24 query heads, six key--value heads, and head dimension 64. It uses the same AdamW $\beta$, $\epsilon$, peak and floor learning rates, and gradient clipping as 337M, with a 2,000-update warmup and a planned 152,588-update cosine schedule. The analyzed checkpoint at 124,000 steps precedes the scheduled learning-rate floor. Both the 1B standard-residual and two-stream runs apply weight decay 0.1 to parameters with at least two dimensions and zero to the remaining parameters. We use 64 calibration and 128 disjoint test sequences
at $t=0.5$. Paired $\Delta$CE measures masked-target loss on identical
inputs and masks; masked/visible result pairs denote the positions modified.
1B 95\% intervals use 5,000 paired bootstrap resamples of sequences. Evidence-transfer intervals resample the 58 sequences contributing the 58 selected pairs, using the same resamples across depths and patch conditions. The 58 selected transfer pairs come from 58 distinct sequences, so sequence-cluster resampling is equivalent to pair resampling for this sample. Matched-step standard-residual / mHC validation CE is
3.27752 / 3.24224, averaged over five mask ratios on a separate split:
a 3.47\% reduction in exponentiated CE (5.71\% at 337M).

\subsection{Early preparation and position-specific preservation}
\label{app:scale-preparation}

\textbf{Early blocks prepare the state that will later be retained.}
The first masked attention update sends 99.67\% of its write gain to
$s_P$ at 1B, versus 97.95\% at 337M. Restoring reference-run post-$m_0$
$s_P$ removes over 99.9\% of the damage from deleting that update in
both models. This restores a stream state after the MLP; the 98.24\%
rescue in Appendix~\ref{app:attn-intervention} restores only the attention
contribution to the MLP input. This rescue identifies $s_P$ as a carrier of early contextual
information needed by subsequent computation.

The state becomes reusable after further early processing at 1B.
Replacing $s_P$ at $a_{16}$ with its earlier state costs
$+0.0442/+0.0164$ CE from $m_1$, but only
$+0.00456/+0.00761$ when taken from $a_8$. At 337M, the state at
$m_0$ already serves at $a_8$ for $+0.0037/+0.0059$ CE.
The declining cost supports a transition from preparation to preservation,
with a later boundary at 1B.

\begin{figure}[htbp]
\centering
\includegraphics[width=\linewidth]{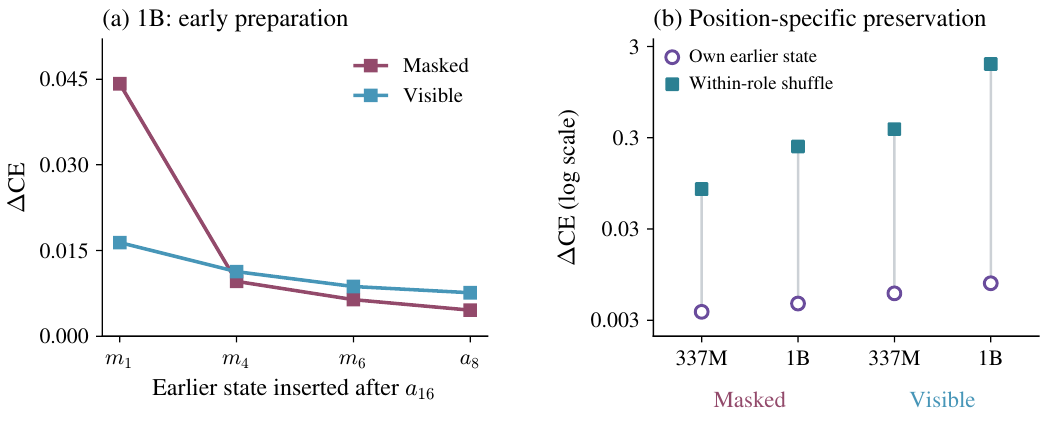}
\caption{\textbf{Preparation leads to position-specific preservation.}
(a) Later early-state anchors are less costly to reuse at 1B.
(b) Own-state reuse versus same-role shuffling, measured across
$m_0\!\to a_8$ (337M) and $a_8\!\to a_{16}$ (1B).
The right axis is logarithmic.}
\label{fig:scale-preparation-preservation}
\end{figure}

\textbf{The preserved information belongs to a particular position.}
Reusing a position's earlier state tests whether its information remains
useful; shuffling within a role tests whether positions' states are
interchangeable. At 1B, the cheap reuse above
contrasts with $+0.2399/+1.9266$ CE for shuffling, using \texttt{SeedSequence([130923, block\_id, depth, stream])}. Reusing the other
stream's earlier state is 35--37 times more costly than reusing $s_P$.
These results support useful position-specific preservation alongside
ongoing computation, as at 337M. Low middle-depth reading and high
same-position cosine agree with this pattern
(Figure~\ref{fig:ar-preservation}).

\subsection{Source influence transfers to targets before late reuse}
\label{app:scale-transfer-reuse}

\textbf{Context reaches the target while position-specific states are preserved.}
The 337M interventions show that source influence accumulates at masked
targets while position-specific information remains available in the
preserved stream. The 1B evidence supports the same combination:
the target state increasingly carries recoverable influence from a visible
source. To measure this, we mask an influential source, then restore reference-run
states at either position. A target patch tests whether its state already
contains useful source influence; a source patch tests whether the remaining
layers can still use restored source information. Recovery is the fraction
of lost correct-token log-probability restored by patching both streams.

Early in 1B (about 6\% depth), restoring the source recovers 92.2\% of
its contribution, while restoring the target recovers only 16.6\%.
At 75\% depth, target recovery reaches 99.8\% and source recovery falls
to 31.2\%; the paired gap is 68.6 percentage points (95\% interval
$[62.0,75.4]$). At the same relative depth, 337M gives 99.6\%/26.3\%
(target/source). The shared trajectory supports accumulation of source
influence at the target alongside preservation of position-specific states
(Figure~\ref{fig:scale-transfer-late}a).

As in Appendix~\ref{app:attn-intervention}, selected pairs require a
source-removal penalty of at least 0.5 log-probability units at a correctly
predicted target. The 1B search yields 58 pairs across 128 sequences.

\begin{figure}[htbp]
\centering
\includegraphics[width=\linewidth]{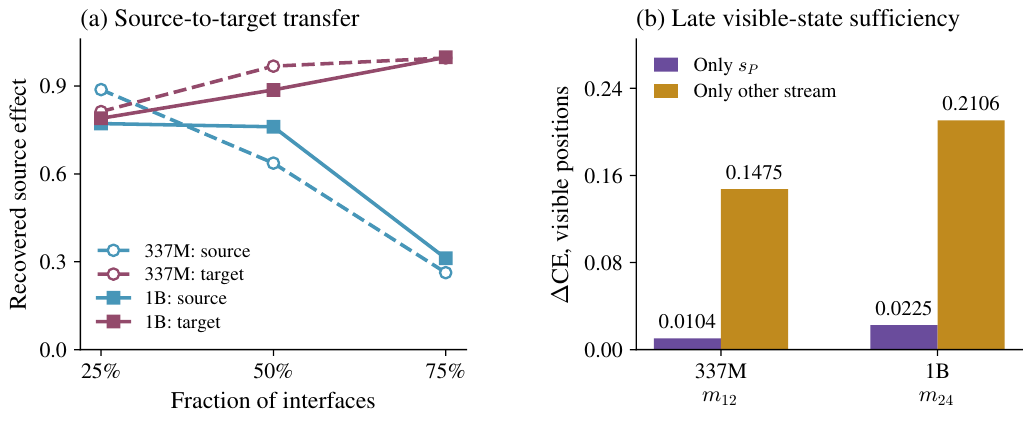}
\caption{\textbf{Source-to-target transfer and late visible-state reuse recur at 1B.} (a) Reference target-state patches recover nearly all
source influence at 75\% depth in both models, while source-state patches
recover much less. Circles/dashes denote 337M; squares/solid lines denote
1B. (b) At $m_{12}$ (337M) and $m_{24}$ (1B), duplicating $s_P$ at visible
positions is much less damaging than duplicating the other stream. The
figures show reported point estimates. The legend's ``Only'' labels denote copying the selected state into both streams.}
\label{fig:scale-transfer-late}
\end{figure}

\textbf{Late computation can rely on the retained visible state.}
At both scales, $s_P$ is read little in the middle and strongly late;
its 1B read share reaches about 87\% after block 24 for both roles.
We test the usefulness of this state by duplicating one stream into both
residual cells at visible positions, leaving masked positions unchanged.
At $m_{24}$ in 1B, duplicating $s_P$ costs $+0.0225$ CE, versus
$+0.2106$ for duplicating the other stream; 337M gives $+0.0104/+0.1475$ at
$m_{12}$. The 1B ordering holds at all six tested boundaries from $m_8$
to $m_{28}$. The retained visible state therefore supplies information
that is sufficient for much of the remaining computation under this
duplication intervention, consistent with its renewed late reading (Figure~\ref{fig:scale-transfer-late}b).

At 1B, patching only $s_P$ or only the other stream at the target recovers
93.9\% or 71.0\% at 75\% depth. Thus, target-side source influence is
available in both streams.

\subsection{Routing reflects the same division between context and prediction}
\label{app:scale-allocation}

\textbf{The interface distinguishes context sources from prediction targets.}
Visible positions provide observed context, while masked positions must
form predictions. As at 337M, this distinction helps explain both which
stream a branch reads and where it writes. Replacing token-specific read
weights with one mean per interface costs $+0.471$ CE at 1B; retaining
separate visible and masked means reduces this to $+0.082$. For normalized write destination, the corresponding costs are $+0.130$ and $+0.036$.
Preserving only the role distinction therefore recovers 82.6\% of read
damage and 72.3\% of destination damage, versus 95.7\% and 98.6\% at
337M (Figure~\ref{fig:scale-role-allocation}a). Role structure remains a
substantial organizing factor.

\textbf{Write volume assignment also adapts to individual tokens and depth.}
Within each role, predictions also depend on how write volumes are
assigned to individual tokens. At 1B visible
positions, shuffling volumes within the role costs $+3.1392$ CE,
versus $+0.1031$ for role averaging; 337M gives $+0.294/+0.124$.
Shuffling preserves the current within-role volume multiset at each interface while changing its assignment to tokens. The CE increase under this distribution-preserving shuffle establishes the
importance of matching write volume to token state within the visible role
at both scales.

The endpoint effects also fit the staged account. At 1B, first-block
role-mean volume replacement mainly affects visible positions
($+0.0598$ CE, versus $+0.00013$ for masked positions); in the last
block it affects masked positions ($+0.00973$, versus zero for visible
positions). The same early-visible / late-masked contrast occurs at
337M (Figure~\ref{fig:scale-role-allocation}b). At both scales, the first-block intervention is much more damaging at visible than masked positions, while masked-token volume remains consequential near the readout. This pattern is consistent with the depth-dependent volume assignment observed across the full 337M network.

Table~\ref{tab:volume-endpoint-coverage-1b} reports acceptance coverage for these 1B endpoint interventions. Coverage is averaged over the two interfaces of the selected block, separately for the masked-only and visible-only runs.

\begin{table}[H]
\centering
\small
\caption{Mean acceptance coverage (\%) for first- and last-block role-mean volume replacement in the 1B model at $r=0.5$. Blocks are indexed from zero.}
\label{tab:volume-endpoint-coverage-1b}
\begin{tabular}{@{}lrr@{}}
\toprule
Block & Masked-only & Visible-only \\
\midrule
0 & 100.00 & 100.00 \\
31 & 86.88 & 99.12 \\
\bottomrule
\end{tabular}
\end{table}

\begin{figure}[htbp]
\centering
\includegraphics[width=\linewidth]{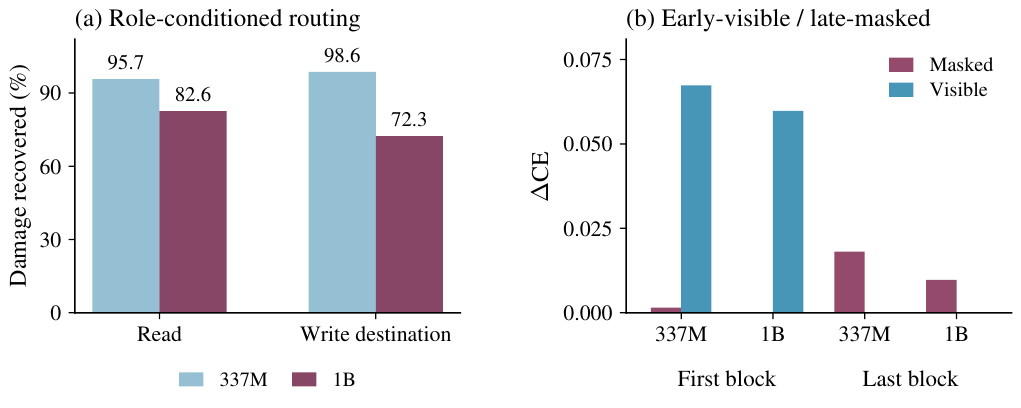}
\caption{\textbf{Role organization and depth-dependent volume assignment recur.}
(a) Fraction of global-mean damage recovered by role means,
$100(1-\Delta\mathrm{CE}_{\mathrm{role}}/\Delta\mathrm{CE}_{\mathrm{global}})$,
computed from reported loss increases.
(b) First/last-block role-mean volume replacements reproduce the
early-visible / late-masked contrast. First- and last-block interventions are shown.}
\label{fig:scale-role-allocation}
\end{figure}

\paragraph{Volume coordinate.} Both scales use the proportional reconstruction and acceptance rules in Appendix~\ref{app:protocol}. Role-mean replacements are applied at feasible positions, while within-role shuffling uses mutually feasible pair exchanges. These interventions preserve each affected token's normalized write destination and keep gains in $[0,2]$; pair exchanges also preserve the complete volume multiset within each sequence, role, and interface. Full-network acceptance coverage for both scales is reported in Table~\ref{tab:volume-coverage}; coverage for the 1B single-block endpoint interventions appears in Table~\ref{tab:volume-endpoint-coverage-1b}.

\end{appendixpart}

\begin{thebibliography}{21}
\providecommand{\natexlab}[1]{#1}
\providecommand{\url}[1]{\texttt{#1}}
\expandafter\ifx\csname urlstyle\endcsname\relax
  \providecommand{\doi}[1]{doi: #1}\else
  \providecommand{\doi}{doi: \begingroup \urlstyle{rm}\Url}\fi

\bibitem[Alimaskina et~al.(2026)Alimaskina, Molodtsov, and
  Beznosikov]{alimaskina2026streamcollapse}
Ekaterina Alimaskina, Gleb Molodtsov, and Aleksandr Beznosikov.
\newblock {Analyzing stream collapse in hyper-connections: From diagnosis to
  mitigation}.
\newblock arXiv preprint arXiv:2606.03483, 2026.
\newblock URL \url{https://arxiv.org/abs/2606.03483}.

\bibitem[Austin et~al.(2021)Austin, Johnson, Ho, Tarlow, and van~den
  Berg]{austin2021structured}
Jacob Austin, Daniel~D. Johnson, Jonathan Ho, Daniel Tarlow, and Rianne van~den
  Berg.
\newblock {Structured denoising diffusion models in discrete state-spaces}.
\newblock In \emph{Advances in Neural Information Processing Systems},
  volume~34, pages 17981--17993, 2021.
\newblock URL \url{https://arxiv.org/abs/2107.03006}.

\bibitem[Bachlechner et~al.(2021)Bachlechner, Majumder, Mao, Cottrell, and
  McAuley]{bachlechner2021rezero}
Thomas Bachlechner, Bodhisattwa~Prasad Majumder, Henry Mao, Gary Cottrell, and
  Julian McAuley.
\newblock {ReZero is all you need: Fast convergence at large depth}.
\newblock In \emph{Proceedings of the Thirty-Seventh Conference on Uncertainty
  in Artificial Intelligence}, volume 161 of \emph{Proceedings of Machine
  Learning Research}, pages 1352--1361. PMLR, 2021.
\newblock URL \url{https://proceedings.mlr.press/v161/bachlechner21a.html}.

\bibitem[Hoogeboom et~al.(2021)Hoogeboom, Nielsen, Jaini, Forr{\'e}, and
  Welling]{hoogeboom2021argmax}
Emiel Hoogeboom, Didrik Nielsen, Priyank Jaini, Patrick Forr{\'e}, and Max
  Welling.
\newblock {Argmax flows and multinomial diffusion: Learning categorical
  distributions}.
\newblock In \emph{Advances in Neural Information Processing Systems},
  volume~34, pages 12454--12465, 2021.
\newblock URL \url{https://arxiv.org/abs/2102.05379}.

\bibitem[{Kimi Team}(2026)]{kimi2026attentionresiduals}
{Kimi Team}.
\newblock {Attention residuals}.
\newblock arXiv preprint arXiv:2603.15031, 2026.
\newblock URL \url{https://arxiv.org/abs/2603.15031}.

\bibitem[Lou et~al.(2024)Lou, Meng, and Ermon]{lou2024discrete}
Aaron Lou, Chenlin Meng, and Stefano Ermon.
\newblock {Discrete diffusion modeling by estimating the ratios of the data
  distribution}.
\newblock In \emph{Proceedings of the 41st International Conference on Machine
  Learning}, 2024.
\newblock URL \url{https://arxiv.org/abs/2310.16834}.

\bibitem[Miele et~al.(2026)Miele, Qin, Carballo-Castro, Deschenaux, and
  Frossard]{miele2026fixedpoint}
Andrea Miele, Yiming Qin, Alba Carballo-Castro, Justin Deschenaux, and Pascal
  Frossard.
\newblock {Fixed-point masked generative modeling}.
\newblock arXiv preprint arXiv:2605.31215, 2026.
\newblock URL \url{https://arxiv.org/abs/2605.31215}.

\bibitem[Nie et~al.(2025)Nie, Zhu, You, Zhang, Ou, Hu, Zhou, Lin, Wen, and
  Li]{nie2025llada}
Shen Nie, Fengqi Zhu, Zebin You, Xiaolu Zhang, Jingyang Ou, Jun Hu, Jun Zhou,
  Yankai Lin, Ji-Rong Wen, and Chongxuan Li.
\newblock Large language diffusion models.
\newblock \emph{Advances in Neural Information Processing Systems},
  38:\penalty0 50608--50646, 2025.

\bibitem[Pagliardini et~al.(2024)Pagliardini, Mohtashami, Fleuret, and
  Jaggi]{pagliardini2024denseformer}
Matteo Pagliardini, Amirkeivan Mohtashami, Fran{\c c}ois Fleuret, and Martin
  Jaggi.
\newblock {DenseFormer: Enhancing information flow in transformers via depth
  weighted averaging}.
\newblock In \emph{Advances in Neural Information Processing Systems},
  volume~37, 2024.
\newblock URL \url{https://arxiv.org/abs/2402.02622}.

\bibitem[Penedo et~al.(2024)Penedo, Kydl{\'i}{\v c}ek, {Ben allal}, Lozhkov,
  Mitchell, Raffel, {Von Werra}, and Wolf]{penedo2024fineweb}
Guilherme Penedo, Hynek Kydl{\'i}{\v c}ek, Loubna {Ben allal}, Anton Lozhkov,
  Margaret Mitchell, Colin Raffel, Leandro {Von Werra}, and Thomas Wolf.
\newblock {The FineWeb datasets: Decanting the web for the finest text data at
  scale}.
\newblock In \emph{Advances in Neural Information Processing Systems},
  volume~37, 2024.
\newblock URL \url{https://arxiv.org/abs/2406.17557}.

\bibitem[Peng et~al.(2026)Peng, Rai, Tseng, Wang, and Wu]{peng2026ablate}
William Peng, Josheev Rai, Kevin Tseng, Siwei Wang, and Sean Wu.
\newblock {Ablate and rescue: A causal analysis of residual stream
  hyper-connections}.
\newblock arXiv preprint arXiv:2603.14833, 2026.
\newblock URL \url{https://arxiv.org/abs/2603.14833}.

\bibitem[Piskorz et~al.(2026)Piskorz, Pinneri, Correia, Alfarra, Garrepalli,
  and Louizos]{piskorz2026masks}
Julianna Piskorz, Cristina Pinneri, Alvaro Correia, Motasem Alfarra, Risheek
  Garrepalli, and Christos Louizos.
\newblock {Masks can be distracting: On context comprehension in diffusion
  language models}.
\newblock In \emph{Proceedings of the 43rd International Conference on Machine
  Learning}, 2026.
\newblock URL \url{https://arxiv.org/abs/2511.21338}.

\bibitem[{Qwen Team}(2024)]{qwen2025qwen25}
{Qwen Team}.
\newblock {Qwen2.5 technical report}.
\newblock arXiv preprint arXiv:2412.15115, 2024.
\newblock URL \url{https://arxiv.org/abs/2412.15115}.

\bibitem[Sahoo et~al.(2024)Sahoo, Arriola, Schiff, Gokaslan, Marroquin, Chiu,
  Rush, and Kuleshov]{sahoo2024simple}
Subham~Sekhar Sahoo, Marianne Arriola, Yair Schiff, Aaron Gokaslan, Edgar
  Marroquin, Justin~T. Chiu, Alexander Rush, and Volodymyr Kuleshov.
\newblock {Simple and effective masked diffusion language models}.
\newblock In \emph{Advances in Neural Information Processing Systems},
  volume~37, pages 130136--130184, 2024.
\newblock URL \url{https://arxiv.org/abs/2406.07524}.

\bibitem[Wang et~al.(2024)Wang, Ma, Dong, Huang, Zhang, and
  Wei]{wang2022deepnet}
Hongyu Wang, Shuming Ma, Li~Dong, Shaohan Huang, Dongdong Zhang, and Furu Wei.
\newblock Deepnet: Scaling transformers to 1,000 layers.
\newblock \emph{IEEE Transactions on Pattern Analysis and Machine
  Intelligence}, 46\penalty0 (10):\penalty0 6761--6774, 2024.

\bibitem[Xie et~al.(2025)Xie, Wei, Cao, Zhao, Deng, Li, Dai, Gao, Chang, Yu,
  Zhao, Zhou, Xu, Zhang, Zeng, Hu, Wang, Yuan, Wang, and Liang]{xie2025mhc}
Zhenda Xie, Yixuan Wei, Huanqi Cao, Chenggang Zhao, Chengqi Deng, Jiashi Li,
  Damai Dai, Huazuo Gao, Jiang Chang, Kuai Yu, Liang Zhao, Shangyan Zhou, Zhean
  Xu, Zhengyan Zhang, Wangding Zeng, Shengding Hu, Yuqing Wang, Jingyang Yuan,
  Lean Wang, and Wenfeng Liang.
\newblock {mHC: Manifold-constrained hyper-connections}.
\newblock arXiv preprint arXiv:2512.24880, 2025.
\newblock URL \url{https://arxiv.org/abs/2512.24880}.

\bibitem[Yan et~al.(2026)Yan, Wang, Wan, Yu, and Tsang]{yan2026starquant}
Xin Yan, Aqiang Wang, Zhenglin Wan, Xingrui Yu, and Ivor~W. Tsang.
\newblock {STaR-Quant: State-time consistent post-training quantization for
  diffusion large language models}.
\newblock arXiv preprint arXiv:2606.04945, 2026.
\newblock URL \url{https://arxiv.org/abs/2606.04945}.

\bibitem[Yang and Gao(2026)]{yang2026mhclite}
Yongyi Yang and Jianyang Gao.
\newblock {mHC-lite: You don{\textquoteright}t need 20 Sinkhorn--Knopp
  iterations}.
\newblock arXiv preprint arXiv:2601.05732, 2026.
\newblock URL \url{https://arxiv.org/abs/2601.05732}.

\bibitem[Ye et~al.(2025)Ye, Xie, Zheng, Gao, Wu, Jiang, Li, and
  Kong]{ye2025dream}
Jiacheng Ye, Zhihui Xie, Lin Zheng, Jiahui Gao, Zirui Wu, Xin Jiang, Zhenguo
  Li, and Lingpeng Kong.
\newblock {Dream 7B: Diffusion large language models}.
\newblock arXiv preprint arXiv:2508.15487, 2025.
\newblock URL \url{https://arxiv.org/abs/2508.15487}.

\bibitem[Zhao et~al.(2026)Zhao, Li, Yu, Guo, and Dong]{zhao2026mhcstreams}
Pengxiang Zhao, Xing Li, Xianzhi Yu, Wei Guo, and Zhenhua Dong.
\newblock {How does mHC use its residual streams? Selective routing and
  near-identity mixing}.
\newblock arXiv preprint arXiv:2609.05309, 2026.
\newblock URL \url{https://arxiv.org/abs/2609.05309}.

\bibitem[Zhu et~al.(2025)Zhu, Huang, Huang, Zeng, Mao, Wu, Min, and
  Zhou]{zhu2025hyperconnections}
Defa Zhu, Hongzhi Huang, Zihao Huang, Yutao Zeng, Yunyao Mao, Banggu Wu, Qiyang
  Min, and Xun Zhou.
\newblock {Hyper-connections}.
\newblock In \emph{International Conference on Learning Representations}, 2025.
\newblock URL \url{https://arxiv.org/abs/2409.19606}.

\end{thebibliography}
\end{document}